\documentclass[10pt,twocolumn,letterpaper]{article}

\usepackage[pagenumbers]{cvpr} 

\usepackage{enumitem}
\usepackage{comment}
\definecolor{cvprblue}{rgb}{0.21,0.49,0.74}
\usepackage[pagebackref,breaklinks,colorlinks,allcolors=cvprblue]{hyperref}

\def\paperID{6047} 
\def\confName{CVPR}
\def\confYear{2025}

\title{FineLIP: Extending CLIP’s Reach via Fine-Grained Alignment with Longer Text Inputs}

\author{Mothilal Asokan\thanks{This work is done when Mothilal Asokan works as an intern at TII.},\ \ Kebin Wu\thanks{Corresponding author}, \ \ Fatima Albreiki \\
Technology Innovation Institute (TII), United Arab Emirates \\
{\tt\small \ {Mothilal.Asokan@mbzuai.ac.ae}}, \ \ 
{\tt\small \{kebin.wu, fatima.albreiki\}@tii.ae}
}

\usepackage{multirow}
\usepackage[super]{nth}
\usepackage{lipsum} 
\usepackage{tablefootnote} 
\usepackage[skip=3pt]{caption}
\usepackage{array}

\begin{document}
\maketitle
\newcommand{\LongCaption}{
\begin{tabular}{l|l|llllll|llllll}
\toprule
 \multicolumn{1}{c}{} & & \multicolumn{6}{c|}{\textbf{Urban1k}} & \multicolumn{6}{c}{\textbf{DOCCI}} \\ \cmidrule{3-14} 
 \multicolumn{1}{c}{} & & \multicolumn{3}{c|}{Image-to-Text} & \multicolumn{3}{c|}{Text-to-Image} & \multicolumn{3}{c|}{Image-to-Text} & \multicolumn{3}{c}{Text-to-Image} \\ \cmidrule{3-14} 
 \multicolumn{1}{c}{} & & \multicolumn{1}{c}{R@1} & \multicolumn{1}{c}{R@5} & \multicolumn{1}{c|}{R@10} & \multicolumn{1}{c}{R@1} & \multicolumn{1}{c}{R@5} & \multicolumn{1}{c|}{R@10} & \multicolumn{1}{c}{R@1} & \multicolumn{1}{c}{R@5} & \multicolumn{1}{c|}{R@10} & \multicolumn{1}{c}{R@1} & \multicolumn{1}{c}{R@5} & \multicolumn{1}{c}{R@10} \\  \midrule 
 \parbox[t]{2mm}{\multirow{6}{*}{\rotatebox[origin=c]{90}{B/16}}} 
& Baseline & 0.859 & 0.969 & \multicolumn{1}{l|}{0.989} & 0.866 & 0.963 & 0.976 & 0.731 & 0.937 & \multicolumn{1}{l|}{0.972} & 0.767 & 0.946 & 0.975 \\
& SPARC \cite{sparc} & 0.854 & 0.963 & \multicolumn{1}{l|}{0.989} & 0.853 & 0.957 & 0.977 & 0.700 & 0.922 & \multicolumn{1}{l|}{0.960} & 0.720 & 0.925 & 0.965 \\
& LAPS \cite{laps} & 0.890 & \textbf{0.987} & \multicolumn{1}{l|}{0.994} & 0.884 & 0.971 & 0.988 & 0.768 & 0.953 & \multicolumn{1}{l|}{0.979} & 0.783 & 0.954 & 0.981 \\
& TULIP \cite{tulip} & 0.881 & - & \multicolumn{1}{l|}{-} & 0.866 & - & - & - & - & \multicolumn{1}{l|}{-} & - & - & - \\
& Long-CLIP \cite{longclip} & 0.789 & - & \multicolumn{1}{l|}{-} & 0.795 & - & - & - & - & \multicolumn{1}{l|}{-} & - & - & - \\
& \textbf{FineLIP (Ours)} & 0.907 & 0.983 & \multicolumn{1}{l|}{\textbf{0.995}} & 0.893 & 0.975 & 0.987 & 0.771 & 0.954 & \multicolumn{1}{l|}{0.980} & 0.795 & 0.958 & \textbf{0.984} \\ 
& \textbf{FineLIP* (Ours)} & \textbf{0.912} & 0.985 & \multicolumn{1}{l|}{\textbf{0.995}} & \textbf{0.900} & \textbf{0.977} & \textbf{0.990} & \textbf{0.781} & \textbf{0.961} & \multicolumn{1}{l|}{\textbf{0.982}} & \textbf{0.800} & \textbf{0.959} & \textbf{0.984} \\ \midrule
\parbox[t]{2mm}{\multirow{6}{*}{\rotatebox[origin=c]{90}{L/14}}} 
& Baseline & 0.892 & 0.985 & \multicolumn{1}{l|}{0.995} & 0.907 & 0.982 & 0.988 & 0.791 & 0.954 & \multicolumn{1}{l|}{0.981} & 0.817 & 0.967 & 0.985 \\
& SPARC \cite{sparc} & 0.772 & 0.940 & \multicolumn{1}{l|}{0.969} & 0.814 & 0.945 & 0.975 & 0.670 & 0.909 & \multicolumn{1}{l|}{0.952} & 0.700 & 0.917 & 0.965 \\
& LAPS \cite{laps} & 0.921 & 0.988 & \multicolumn{1}{l|}{0.994} & 0.916 & 0.981 & 0.988 & 0.818 & 0.965 & \multicolumn{1}{l|}{0.985} & 0.818 & 0.965 & 0.985 \\
& TULIP \cite{tulip} & 0.901 & - & \multicolumn{1}{l|}{-} & 0.911 & - & - & - & - & \multicolumn{1}{l|}{-} & - & - & - \\
& Long-CLIP \cite{longclip} & 0.827 & - & \multicolumn{1}{l|}{-} & 0.861 & - & - & - & - & \multicolumn{1}{l|}{-} & - & - & - \\
& \textbf{FineLIP (Ours)} & 0.932 & 0.988 & \multicolumn{1}{l|}{\textbf{0.996}} & 0.930 & 0.984 & 0.994 & 0.822 & 0.967 & \multicolumn{1}{l|}{0.985} & 0.831 & 0.971 & 0.990 \\ 
& \textbf{FineLIP* (Ours)} & \textbf{0.945} & \textbf{0.993} & \multicolumn{1}{l|}{\textbf{0.996}} & \textbf{0.939} & \textbf{0.987} & \textbf{0.995} & \textbf{0.837} & \textbf{0.972} & \multicolumn{1}{l|}{\textbf{0.989}} & \textbf{0.844} & \textbf{0.974} & \textbf{0.991} \\ \midrule 
\parbox[t]{2mm}{\multirow{4}{*}{\rotatebox[origin=c]{90}{bigG/14}}} 
& Baseline & 0.908 & 0.986 & \multicolumn{1}{l|}{\textbf{0.996}} & 0.911 & 0.977 & 0.985 & 0.813 & 0.965 & \multicolumn{1}{l|}{0.984} & 0.842 & 0.972 & 0.987 \\
& LAPS \cite{laps} & \textbf{0.932} & 0.988 & \multicolumn{1}{l|}{\textbf{0.996}} & 0.928 & 0.984 & 0.991 & 0.832 & 0.968 & \multicolumn{1}{l|}{0.988} & 0.848 & 0.975 & 0.990 \\
& \textbf{FineLIP (Ours)} & 0.924 & 0.991 & \multicolumn{1}{l|}{0.995} & 0.933 & 0.985 & 0.991 & 0.836 & 0.972 & \multicolumn{1}{l|}{0.988} & 0.853 & 0.975 & 0.989 \\ 
& \textbf{FineLIP* (Ours)} & \textbf{0.932} & \textbf{0.992} & \multicolumn{1}{l|}{\textbf{0.996}} & \textbf{0.941} & \textbf{0.986} & \textbf{0.994} & \textbf{0.845} & \textbf{0.974} & \multicolumn{1}{l|}{\textbf{0.990}} & \textbf{0.860} & \textbf{0.978} & \textbf{0.992} \\ 
\bottomrule 
\end{tabular}
}

\newcommand{\fid}{
\small
\begin{tabular}{l|l|ll}
\toprule
\multicolumn{1}{c|}{\multirow{2}{*}{}} & \multicolumn{2}{c}{FID $\downarrow$} \\ \cmidrule{2-3}
\multicolumn{1}{c|}{L/14} & \multicolumn{1}{c|}{Urban1k} & \multicolumn{1}{c}{DOCCI} \\ \midrule
Baseline & \multicolumn{1}{l|}{29.535} & 16.884 \\
SPARC \cite{sparc} & \multicolumn{1}{l|}{31.604} & 17.241 \\
LAPS \cite{laps} & \multicolumn{1}{l|}{\textbf{26.743}} & 15.426 \\
\textbf{FineLIP (Ours)} & \multicolumn{1}{l|}{27.261} & \textbf{15.410} \\ \midrule
\end{tabular}
}

\newcommand{\InitA}{
\begin{tabular}{l|llllll|llllll}
\toprule 
 & \multicolumn{6}{c|}{\textbf{Urban1k}} & \multicolumn{6}{c}{\textbf{DOCCI}} \\ \cmidrule{2-13} 
\multicolumn{1}{c|}{\textbf{B/16}} & \multicolumn{3}{c|}{Image-to-Text} & \multicolumn{3}{c|}{Text-to-Image} & \multicolumn{3}{c|}{Image-to-Text} & \multicolumn{3}{c}{Text-to-Image} \\ \cmidrule{2-13} 
 & \multicolumn{1}{c}{R@1} & \multicolumn{1}{c}{R@5} & \multicolumn{1}{c|}{R@10} & \multicolumn{1}{c}{R@1} & \multicolumn{1}{c}{R@5} & \multicolumn{1}{c|}{R@10} & \multicolumn{1}{c}{R@1} & \multicolumn{1}{c}{R@5} & \multicolumn{1}{c|}{R@10} & \multicolumn{1}{c}{R@1} & \multicolumn{1}{c}{R@5} & \multicolumn{1}{c}{R@10} \\ \midrule 
Baseline$\ddag$ & 0.298 & 0.582 & \multicolumn{1}{l|}{0.700} & 0.306 & 0.559 & 0.667 & 0.155 & 0.344 & \multicolumn{1}{l|}{0.448} & 0.115 & 0.276 & 0.372 \\
Pretrained-CLIP$\mathsection$ & 0.684 & 0.888 & \multicolumn{1}{l|}{0.939} & 0.559 & 0.796 & 0.863 & 0.658 & 0.892 & \multicolumn{1}{l|}{0.935} & 0.631 & 0.871 & 0.923 \\
Baseline & 0.859 & 0.969 & \multicolumn{1}{l|}{0.989} & 0.866 & 0.963 & 0.976 & 0.731 & 0.937 & \multicolumn{1}{l|}{0.972} & 0.767 & 0.946 & 0.975 \\
\textbf{FineLIP (Ours)} & 0.907 & 0.983 & \multicolumn{1}{l|}{0.995} & 0.893 & 0.975 & 0.987 & 0.771 & 0.954 & \multicolumn{1}{l|}{0.980} & 0.795 & 0.958 & 0.984 \\ \bottomrule 
\end{tabular}
}

\newcommand{\InitB}{
\begin{tabular}{l|llll|llll}
\toprule
 & \multicolumn{4}{c|}{\textbf{Urban1k}} & \multicolumn{4}{c}{\textbf{DOCCI}} \\ \cmidrule{2-9}
\multicolumn{1}{c|}{\textbf{B/16}} & \multicolumn{2}{c|}{Image-to-Text} & \multicolumn{2}{c|}{Text-to-Image} & \multicolumn{2}{c|}{Image-to-Text} & \multicolumn{2}{c}{Text-to-Image} \\ \cmidrule{2-9}
 & \multicolumn{1}{c}{R@1} & \multicolumn{1}{c|}{R@5} & \multicolumn{1}{c}{R@1} & \multicolumn{1}{c|}{R@5} & \multicolumn{1}{c}{R@1} & \multicolumn{1}{c|}{R@5} & \multicolumn{1}{c}{R@1} & \multicolumn{1}{c}{R@5} \\ \midrule
Baseline$\ddag$ & 0.298 & \multicolumn{1}{l|}{0.582} & 0.306 & 0.559 & 0.155 & \multicolumn{1}{l|}{0.344} & 0.115 & 0.276 \\
Pretrained-CLIP$\mathsection$ & 0.684 & \multicolumn{1}{l|}{0.888} & 0.559 & 0.796 & 0.658 & \multicolumn{1}{l|}{0.892} & 0.631 & 0.871 \\
Baseline & 0.859 & \multicolumn{1}{l|}{0.969} & 0.866 & 0.963 & 0.731 & \multicolumn{1}{l|}{0.937} & 0.767 & 0.946 \\
\bottomrule 
\end{tabular}
}

\newcommand{\atrmA}{
\begin{tabular}{c|llllll|llllll}
\toprule 
 \multirow{2}{*}{\textbf{\begin{tabular}[c]{@{}c@{}}Different\\ Settings\end{tabular}}} & \multicolumn{6}{c|}{\textbf{Urban1k}} & \multicolumn{6}{c}{\textbf{DOCCI}} \\ \cmidrule{2-13} 
 & \multicolumn{3}{c|}{Image-to-Text} & \multicolumn{3}{c|}{Text-to-Image} & \multicolumn{3}{c|}{Image-to-Text} & \multicolumn{3}{c}{Text-to-Image} \\ \cmidrule{2-13} 
 & \multicolumn{1}{c}{R@1} & \multicolumn{1}{c}{R@5} & \multicolumn{1}{c|}{R@10} & \multicolumn{1}{c}{R@1} & \multicolumn{1}{c}{R@5} & \multicolumn{1}{c|}{R@10} & \multicolumn{1}{c}{R@1} & \multicolumn{1}{c}{R@5} & \multicolumn{1}{c|}{R@10} & \multicolumn{1}{c}{R@1} & \multicolumn{1}{c}{R@5} & \multicolumn{1}{c}{R@10} \\ \midrule 
CLIM (only) & 0.854 & 0.967 & \multicolumn{1}{l|}{0.987} & 0.782 & 0.940 & 0.965 & 0.721 & 0.928 & \multicolumn{1}{l|}{0.965} & 0.701 & 0.915 & 0.959 \\
ATRM (image) + CLIM & 0.893 & 0.985 & \multicolumn{1}{l|}{0.992} & 0.890 & 0.972 & 0.984 & 0.767 & 0.953 & \multicolumn{1}{l|}{0.979} & 0.786 & 0.953 & 0.980 \\
ATRM (text) + CLIM & 0.887 & 0.975 & \multicolumn{1}{l|}{0.987} & 0.827 & 0.964 & 0.983 & 0.767 & 0.955 & \multicolumn{1}{l|}{0.979} & 0.742 & 0.936 & 0.965 \\
ATRM (both) + CLIM & 0.907 & 0.983 & \multicolumn{1}{l|}{0.995} & 0.893 & 0.975 & 0.987 & 0.771 & 0.954 & \multicolumn{1}{l|}{0.980} & 0.795 & 0.958 & 0.984 \\ \bottomrule 
\end{tabular}
}

\newcommand{\atrmB}{
\begin{tabular}{c|llll|llll}
\toprule
 \multirow{2}{*}{\textbf{\begin{tabular}[c]{@{}c@{}}Different\\ Settings\end{tabular}}} & \multicolumn{4}{c|}{\textbf{Urban1k}} & \multicolumn{4}{c}{\textbf{DOCCI}} \\ \cmidrule{2-9}
 & \multicolumn{2}{c|}{Image-to-Text} & \multicolumn{2}{c|}{Text-to-Image} & \multicolumn{2}{c|}{Image-to-Text} & \multicolumn{2}{c}{Text-to-Image} \\ \cmidrule{2-9} 
 & \multicolumn{1}{c}{R@1} & \multicolumn{1}{c|}{R@5} & \multicolumn{1}{c}{R@1} & \multicolumn{1}{c|}{R@5} & \multicolumn{1}{c}{R@1} & \multicolumn{1}{c|}{R@5} & \multicolumn{1}{c}{R@1} & \multicolumn{1}{c}{R@5} \\ \midrule
CLIM (only) & 0.854 & \multicolumn{1}{l|}{0.967} & 0.782 & 0.940 & 0.721 & \multicolumn{1}{l|}{0.928} & 0.701 & 0.915 \\
ATRM (image) + CLIM & 0.893 & \multicolumn{1}{l|}{0.985} & 0.890 & 0.972 & 0.767 & \multicolumn{1}{l|}{0.953} & 0.786 & 0.953 \\
ATRM (text) + CLIM & 0.887 & \multicolumn{1}{l|}{0.975} & 0.827 & 0.964 & 0.767 & \multicolumn{1}{l|}{0.955} & 0.742 & 0.936 \\
ATRM (both) + CLIM & 0.907 & \multicolumn{1}{l|}{0.983} & 0.893 & 0.975 & 0.771 & \multicolumn{1}{l|}{0.954} & 0.795 & 0.958 \\ \bottomrule
\end{tabular}
}

\newcommand{\AggregationRatioA}{
\begin{tabular}{c|llllll|llllll}
\toprule 
 \multirow{3}{*}{\textbf{\begin{tabular}[c]{@{}c@{}}Aggre- \\gation\\ ratio\end{tabular}}} & \multicolumn{6}{c|}{\textbf{Urban1k}} & \multicolumn{6}{c}{\textbf{DOCCI}} \\ \cmidrule{2-13} 
 & \multicolumn{3}{c|}{Image-to-Text} & \multicolumn{3}{c|}{Text-to-Image} & \multicolumn{3}{c|}{Image-to-Text} & \multicolumn{3}{c}{Text-to-Image} \\ \cmidrule{2-13} 
 & \multicolumn{1}{c}{R@1} & \multicolumn{1}{c}{R@5} & \multicolumn{1}{c|}{R@10} & \multicolumn{1}{c}{R@1} & \multicolumn{1}{c}{R@5} & \multicolumn{1}{c|}{R@10} & \multicolumn{1}{c}{R@1} & \multicolumn{1}{c}{R@5} & \multicolumn{1}{c|}{R@10} & \multicolumn{1}{c}{R@1} & \multicolumn{1}{c}{R@5} & \multicolumn{1}{c}{R@10} \\ \midrule 
0.4 & 0.899 & 0.980 & \multicolumn{1}{l|}{0.992} & 0.891 & 0.974 & 0.981 & 0.768 & 0.954 & \multicolumn{1}{l|}{0.979} & 0.785 & 0.957 & 0.982 \\
\textbf{0.2} & 0.907 & 0.983 & \multicolumn{1}{l|}{0.995} & 0.893 & 0.975 & 0.987 & 0.771 & 0.954 & \multicolumn{1}{l|}{0.980} & 0.795 & 0.958 & 0.984 \\
0.1 & 0.905 & 0.984 & \multicolumn{1}{l|}{0.995} & 0.883 & 0.978 & 0.988 & 0.776 & 0.958 & \multicolumn{1}{l|}{0.982} & 0.790 & 0.959 & 0.983 \\ \bottomrule 
\end{tabular}
}

\newcommand{\AggregationRatioB}{
\begin{tabular}{c|llll|llll}
\toprule 
 \multirow{3}{*}{\textbf{\begin{tabular}[c]{@{}c@{}}Aggre- \\gation\\ ratio\end{tabular}}} & \multicolumn{4}{c|}{\textbf{Urban1k}} & \multicolumn{4}{c}{\textbf{DOCCI}} \\ \cmidrule{2-9}
 & \multicolumn{2}{c|}{Image-to-Text} & \multicolumn{2}{c|}{Text-to-Image} & \multicolumn{2}{c|}{Image-to-Text} & \multicolumn{2}{c}{Text-to-Image} \\ \cmidrule{2-9} 
 & \multicolumn{1}{c}{R@1} & \multicolumn{1}{c|}{R@5} & \multicolumn{1}{c}{R@1} & \multicolumn{1}{c|}{R@5} & \multicolumn{1}{c}{R@1} & \multicolumn{1}{c|}{R@5} & \multicolumn{1}{c}{R@1} & \multicolumn{1}{c}{R@5} \\ \midrule
0.4 & 0.899 & \multicolumn{1}{l|}{0.980} & 0.891 & 0.974 & 0.768 & \multicolumn{1}{l|}{0.954} & 0.785 & 0.957 \\
\textbf{0.2} & 0.907 & \multicolumn{1}{l|}{0.983} & 0.893 & 0.975 & 0.771 & \multicolumn{1}{l|}{0.954} & 0.795 & 0.958 \\
0.1 & 0.905 & \multicolumn{1}{l|}{0.984} & 0.883 & 0.978 & 0.776 & \multicolumn{1}{l|}{0.958} & 0.790 & 0.959 \\ \bottomrule 
\end{tabular}
}

\newcommand{\GlobalFeatureA}{
\begin{tabular}{l|llllll|llllll}
\toprule 
 & \multicolumn{6}{c|}{\textbf{Urban1k}} & \multicolumn{6}{c}{\textbf{DOCCI}} \\ \cmidrule{2-13} 
 \textbf{B/16} & \multicolumn{3}{c|}{Image-to-Text} & \multicolumn{3}{c|}{Text-to-Image} & \multicolumn{3}{c|}{Image-to-Text} & \multicolumn{3}{c}{Text-to-Image} \\ \cmidrule{2-13} 
 & \multicolumn{1}{c}{R@1} & \multicolumn{1}{c}{R@5} & \multicolumn{1}{c|}{R@10} & \multicolumn{1}{c}{R@1} & \multicolumn{1}{c}{R@5} & \multicolumn{1}{c|}{R@10} & \multicolumn{1}{c}{R@1} & \multicolumn{1}{c}{R@5} & \multicolumn{1}{c|}{R@10} & \multicolumn{1}{c}{R@1} & \multicolumn{1}{c}{R@5} & \multicolumn{1}{c}{R@10} \\ \midrule 
No Global Tokens & 0.894 & 0.982 & \multicolumn{1}{l|}{0.991} & 0.893 & 0.973 & 0.983 & 0.760 & 0.948 & \multicolumn{1}{l|}{0.976} & 0.773 & 0.949 & 0.976 \\
\textbf{FineLIP (Ours)} & 0.907 & 0.983 & \multicolumn{1}{l|}{0.995} & 0.893 & 0.975 & 0.987 & 0.771 & 0.954 & \multicolumn{1}{l|}{0.980} & 0.795 & 0.958 & 0.984 \\ \bottomrule 
\end{tabular}
}

\newcommand{\GlobalFeatureB}{
\begin{tabular}{l|llll|llll}
\toprule 
 & \multicolumn{4}{c|}{\textbf{Urban1k}} & \multicolumn{4}{c}{\textbf{DOCCI}} \\ \cmidrule{2-9} 
 \textbf{B/16} & \multicolumn{2}{c|}{Image-to-Text} & \multicolumn{2}{c|}{Text-to-Image} & \multicolumn{2}{c|}{Image-to-Text} & \multicolumn{2}{c}{Text-to-Image} \\ \cmidrule{2-9} 
 & \multicolumn{1}{c}{R@1} & \multicolumn{1}{c|}{R@5} & \multicolumn{1}{c}{R@1} & \multicolumn{1}{c|}{R@5} & \multicolumn{1}{c}{R@1} & \multicolumn{1}{c|}{R@5} & \multicolumn{1}{c}{R@1} & \multicolumn{1}{c}{R@5} \\ \midrule 
No Global Tokens & 0.894 & \multicolumn{1}{l|}{0.982} & 0.893 & 0.973 & 0.760 & \multicolumn{1}{l|}{0.948} & 0.773 & 0.949 \\
\textbf{FineLIP (Ours)} & 0.907 & \multicolumn{1}{l|}{0.983} & 0.893 & 0.975 & 0.771 & \multicolumn{1}{l|}{0.954} & 0.795 & 0.958 \\ \bottomrule 
\end{tabular}
}

\newcommand{\ablation}{
\begin{tabular}{c|llll|llll}
\multicolumn{9}{c}{A: Impact of Initialization and Positional Embedding Stretching.} \\ \toprule
 & \multicolumn{4}{c|}{\textbf{Urban1k}} & \multicolumn{4}{c}{\textbf{DOCCI}} \\ \cmidrule{2-9}
\multicolumn{1}{c|}{\textbf{Different Settings}} & \multicolumn{2}{c|}{Image-to-Text} & \multicolumn{2}{c|}{Text-to-Image} & \multicolumn{2}{c|}{Image-to-Text} & \multicolumn{2}{c}{Text-to-Image} \\ \cmidrule{2-9} 
 & \multicolumn{1}{c}{R@1} & \multicolumn{1}{c|}{R@5} & \multicolumn{1}{c}{R@1} & \multicolumn{1}{c|}{R@5} & \multicolumn{1}{c}{R@1} & \multicolumn{1}{c|}{R@5} & \multicolumn{1}{c}{R@1} & \multicolumn{1}{c}{R@5} \\ \midrule
Baseline\_rand\_init & 0.298 & \multicolumn{1}{l|}{0.582} & 0.306 & 0.559 & 0.155 & \multicolumn{1}{l|}{0.344} & 0.115 & 0.276 \\
Pretrained-CLIP\tablefootnote{\href{https://openaipublic.azureedge.net/clip/models/5806e77cd80f8b59890b7e101eabd078d9fb84e6937f9e85e4ecb61988df416f/ViT-B-16.pt}{\nolinkurl{CLIP-ViT-B/16}}} & 0.684 & \multicolumn{1}{l|}{0.888} & 0.559 & 0.796 & 0.658 & \multicolumn{1}{l|}{0.892} & 0.631 & 0.871 \\
Baseline\_nPE & 0.697 & \multicolumn{1}{l|}{0.907} & 0.733 & 0.911 & 0.611 & \multicolumn{1}{l|}{0.873} & 0.666 & 0.891 \\
Baseline\_Lv \cite{lit} 
& 0.775 & \multicolumn{1}{l|}{0.950} & 0.755 & 0.931 & 0.650 & \multicolumn{1}{l|}{0.902} & 0.679 & 0.905 \\
Baseline & \textbf{0.859} & \multicolumn{1}{l|}{\textbf{0.969}} & \textbf{0.866} & \textbf{0.963} & \textbf{0.731} & \multicolumn{1}{l|}{\textbf{0.937}} & \textbf{0.767} & \textbf{0.946} \\
\bottomrule 
\multicolumn{9}{c}{B: Impact of ATRM in our model.} \\ \toprule
\multicolumn{1}{c|}{\textbf{Different Settings}} & \multicolumn{1}{c}{R@1} & \multicolumn{1}{c|}{R@5} & \multicolumn{1}{c}{R@1} & \multicolumn{1}{c|}{R@5} & \multicolumn{1}{c}{R@1} & \multicolumn{1}{c|}{R@5} & \multicolumn{1}{c}{R@1} & \multicolumn{1}{c}{R@5} \\ \midrule
CLIM (only) & 0.854 & \multicolumn{1}{l|}{0.967} & 0.782 & 0.940 & 0.721 & \multicolumn{1}{l|}{0.928} & 0.701 & 0.915 \\
ATRM (image) + CLIM & 0.893 & \multicolumn{1}{l|}{\textbf{0.985}} & 0.890 & 0.972 & 0.767 & \multicolumn{1}{l|}{0.953} & 0.786 & 0.953 \\
ATRM (text) + CLIM & 0.887 & \multicolumn{1}{l|}{0.975} & 0.827 & 0.964 & 0.767 & \multicolumn{1}{l|}{\textbf{0.955}} & 0.742 & 0.936 \\
ATRM (both) + CLIM & \textbf{0.907} & \multicolumn{1}{l|}{0.983} & \textbf{0.893} & \textbf{0.975} & \textbf{0.771} & \multicolumn{1}{l|}{0.954} & \textbf{0.795} & \textbf{0.958} \\ \bottomrule 
\multicolumn{9}{c}{C: Effect of different aggregation ratio in ATRM.} \\ \toprule
\multicolumn{1}{c|}{\textbf{Aggregation Ratio}} & \multicolumn{1}{c}{R@1} & \multicolumn{1}{c|}{R@5} & \multicolumn{1}{c}{R@1} & \multicolumn{1}{c|}{R@5} & \multicolumn{1}{c}{R@1} & \multicolumn{1}{c|}{R@5} & \multicolumn{1}{c}{R@1} & \multicolumn{1}{c}{R@5} \\ \midrule
0.4 & 0.899 & \multicolumn{1}{l|}{0.980} & 0.891 & 0.974 & 0.768 & \multicolumn{1}{l|}{0.954} & 0.785 & 0.957 \\
\textbf{0.2} & \textbf{0.907} & \multicolumn{1}{l|}{0.983} & \textbf{0.893} & 0.975 & 0.771 & \multicolumn{1}{l|}{0.954} & \textbf{0.795} & 0.958 \\
0.1 & 0.905 & \multicolumn{1}{l|}{\textbf{0.984}} & 0.883 & \textbf{0.978} & \textbf{0.776} & \multicolumn{1}{l|}{\textbf{0.958}} & 0.790 & \textbf{0.959} \\ \bottomrule 
\multicolumn{9}{c}{D: Importance of retaining global features in alignment.} \\ \toprule
\textbf{B/16} & \multicolumn{1}{c}{R@1} & \multicolumn{1}{c|}{R@5} & \multicolumn{1}{c}{R@1} & \multicolumn{1}{c|}{R@5} & \multicolumn{1}{c}{R@1} & \multicolumn{1}{c|}{R@5} & \multicolumn{1}{c}{R@1} & \multicolumn{1}{c}{R@5} \\ \midrule 
No Global Tokens & 0.894 & \multicolumn{1}{l|}{0.982} & \textbf{0.893} & 0.973 & 0.760 & \multicolumn{1}{l|}{0.948} & 0.773 & 0.949 \\
\textbf{FineLIP (Ours)} & \textbf{0.907} & \multicolumn{1}{l|}{\textbf{0.983}} & \textbf{0.893} & \textbf{0.975} & \textbf{0.771} & \multicolumn{1}{l|}{\textbf{0.954}} & \textbf{0.795} & \textbf{0.958} \\ \bottomrule 
\end{tabular}
}

\newcommand{\additionalExp}{
\begin{tabular}{c|llll|llll}
\toprule
 & \multicolumn{4}{c|}{\textbf{Long-DCI [30]}} & \multicolumn{4}{c}{\textbf{IIW [8]}} \\ \cmidrule{2-9}
\multicolumn{1}{c|}{\textbf{L/14}} & \multicolumn{2}{c|}{Image-to-Text} & \multicolumn{2}{c|}{Text-to-Image} & \multicolumn{2}{c|}{Image-to-Text} & \multicolumn{2}{c}{Text-to-Image} \\ \cmidrule{2-9} 
 & \multicolumn{1}{c}{R@1} & \multicolumn{1}{c|}{R@5} & \multicolumn{1}{c}{R@1} & \multicolumn{1}{c|}{R@5} & \multicolumn{1}{c}{R@1} & \multicolumn{1}{c|}{R@5} & \multicolumn{1}{c}{R@1} & \multicolumn{1}{c}{R@5} \\ \midrule
Baseline & 0.568 & \multicolumn{1}{l|}{0.763} & 0.585 & 0.754 & 0.968 & \multicolumn{1}{l|}{1.000} & 0.978 & 1.000 \\
SPARC [2] & 0.428 & \multicolumn{1}{l|}{0.638} & 0.465 & 0.669 & 0.903 & \multicolumn{1}{l|}{0.988} & 0.933 & 0.998 \\
LAPS [6] & 0.604 & \multicolumn{1}{l|}{0.778} & 0.592 & 0.758 & 0.985 & \multicolumn{1}{l|}{1.000} & 0.983 & 1.000 \\
\textbf{FineLIP (Ours)} & 0.608 & \multicolumn{1}{l|}{0.780} & 0.607 & 0.767 & 0.993 & \multicolumn{1}{l|}{1.000} & 0.985 & 1.000 \\ \bottomrule 
\end{tabular}
}

\newcommand{\rebuttal}{
\begin{tabular}{c|llll|llll}
\multicolumn{9}{c}{A: Long-context retrieval.} \\ \toprule
 & \multicolumn{4}{c|}{\textbf{Long-DCI [30]}} & \multicolumn{4}{c}{\textbf{IIW [8]}} \\ \cmidrule{2-9}
\multicolumn{1}{c|}{\textbf{L/14}} & \multicolumn{2}{c|}{Image-to-Text} & \multicolumn{2}{c|}{Text-to-Image} & \multicolumn{2}{c|}{Image-to-Text} & \multicolumn{2}{c}{Text-to-Image} \\ \cmidrule{2-9} 
 & \multicolumn{1}{c}{R@1} & \multicolumn{1}{c|}{R@5} & \multicolumn{1}{c}{R@1} & \multicolumn{1}{c|}{R@5} & \multicolumn{1}{c}{R@1} & \multicolumn{1}{c|}{R@5} & \multicolumn{1}{c}{R@1} & \multicolumn{1}{c}{R@5} \\ \midrule
Baseline & 0.568 & \multicolumn{1}{l|}{0.763} & 0.585 & 0.754 & 0.968 & \multicolumn{1}{l|}{1.000} & 0.978 & 1.000 \\
SPARC [2] & 0.428 & \multicolumn{1}{l|}{0.638} & 0.465 & 0.669 & 0.903 & \multicolumn{1}{l|}{0.988} & 0.933 & 0.998 \\
LAPS [6] & 0.604 & \multicolumn{1}{l|}{0.778} & 0.592 & 0.758 & 0.985 & \multicolumn{1}{l|}{1.000} & 0.983 & 1.000 \\
\textbf{FineLIP (Ours)} & 0.608 & \multicolumn{1}{l|}{0.780} & 0.607 & 0.767 & 0.993 & \multicolumn{1}{l|}{1.000} & 0.985 & 1.000 \\ 
\bottomrule 
\multicolumn{9}{c}{B: Ablations.} \\ \toprule
\multicolumn{1}{c|}{\textbf{L/14}} & \multicolumn{4}{c|}{\textbf{Urban1k [33]}} & \multicolumn{4}{c}{\textbf{DOCCI [19]}} \\ \cmidrule{2-9}
\multicolumn{1}{c|}{\textbf{Different Settings}} & \multicolumn{1}{c}{R@1} & \multicolumn{1}{c|}{R@5} & \multicolumn{1}{c}{R@1} & \multicolumn{1}{c|}{R@5} & \multicolumn{1}{c}{R@1} & \multicolumn{1}{c|}{R@5} & \multicolumn{1}{c}{R@1} & \multicolumn{1}{c}{R@5} \\ \midrule
Baseline\_Lv 
& 0.775 & \multicolumn{1}{l|}{0.950} & 0.755 & 0.931 & 0.650 & \multicolumn{1}{l|}{0.902} & 0.679 & 0.905 \\
Baseline\_nPE & 0.697 & \multicolumn{1}{l|}{0.907} & 0.733 & 0.911 & 0.611 & \multicolumn{1}{l|}{0.873} & 0.666 & 0.891 \\
ATRM (both) + FILIP & 0.315 & \multicolumn{1}{l|}{0.525} & 0.436 & 0.657 & 0.354 & \multicolumn{1}{l|}{0.620} & 0.426 & 0.694 \\ \bottomrule 
\end{tabular}
}

\newcommand{\ShortCaptionA}{
\begin{tabular}{l|l|llllll|llllll}
\toprule 
 \multicolumn{1}{c}{} & & \multicolumn{6}{c|}{\textbf{Flickr30k}} & \multicolumn{6}{c}{\textbf{COCO}} \\ \cmidrule{3-14} 
 \multicolumn{1}{c}{} & & \multicolumn{3}{c|}{Image-to-Text} & \multicolumn{3}{c|}{Text-to-Image} & \multicolumn{3}{c|}{Image-to-Text} & \multicolumn{3}{c}{Text-to-Image} \\ \cmidrule{3-14} 
 \multicolumn{1}{c}{} & & \multicolumn{1}{c}{R@1} & \multicolumn{1}{c}{R@5} & \multicolumn{1}{c|}{R@10} & \multicolumn{1}{c}{R@1} & \multicolumn{1}{c}{R@5} & \multicolumn{1}{c|}{R@10} & \multicolumn{1}{c}{R@1} & \multicolumn{1}{c}{R@5} & \multicolumn{1}{c|}{R@10} & \multicolumn{1}{c}{R@1} & \multicolumn{1}{c}{R@5} & \multicolumn{1}{c}{R@10} \\ \midrule 
\parbox[t]{2mm}{\multirow{4}{*}{\rotatebox[origin=c]{90}{B/16}}} 
& Pretrained-CLIP & 0.441 & 0.682 & \multicolumn{1}{l|}{0.771} & 0.248 & 0.451 & 0.546 & 0.517 & 0.768 & \multicolumn{1}{l|}{0.843} & 0.327 & 0.578 & 0.682 \\
& Baseline & 0.470 & 0.719 & \multicolumn{1}{l|}{0.805} & 0.334 & 0.564 & \textbf{0.659} & 0.553 & 0.800 & \multicolumn{1}{l|}{0.869} & \textbf{0.409} & \textbf{0.669} & \textbf{0.766} \\
& TULIP \cite{tulip} & 0.461 & 0.708 & \multicolumn{1}{l|}{-} & \textbf{0.352} & \textbf{0.572} & - & 0.568 & 0.803 & \multicolumn{1}{l|}{-} & 0.407 & 0.661 & - \\
& \textbf{FineLIP (Ours)} & \textbf{0.528} & \textbf{0.753} & \multicolumn{1}{l|}{\textbf{0.829}} & 0.341 & 0.567 & 0.658 & \textbf{0.587} & \textbf{0.814} & \multicolumn{1}{l|}{\textbf{0.882}} & 0.404 & 0.662 & 0.760 \\ \midrule 
\parbox[t]{2mm}{\multirow{4}{*}{\rotatebox[origin=c]{90}{L/14}}} 
& Pretrained-CLIP & 0.485 & 0.727 & \multicolumn{1}{l|}{0.809} & 0.280 & 0.493 & 0.587 & 0.560 & 0.795 & \multicolumn{1}{l|}{0.869} & 0.353 & 0.600 & 0.701 \\
& Baseline & 0.544 & 0.788 & \multicolumn{1}{l|}{0.862} & 0.419 & \textbf{0.653} & \textbf{0.739} & 0.608 & 0.836 & \multicolumn{1}{l|}{0.899} & \textbf{0.472} & \textbf{0.721} & \textbf{0.809} \\
& TULIP \cite{tulip} & 0.567 & 0.795 & \multicolumn{1}{l|}{-} & 0.416 & 0.643 & - & 0.626 & 0.847 & \multicolumn{1}{l|}{-} & 0.461 & 0.711 & - \\
& \textbf{FineLIP (Ours)} & \textbf{0.622} & \textbf{0.828} & \multicolumn{1}{l|}{\textbf{0.888}} & \textbf{0.424} & 0.652 & 0.736 & \textbf{0.634} & \textbf{0.848} & \multicolumn{1}{l|}{\textbf{0.910}} & 0.462 & 0.712 & 0.801 \\ \midrule 
\end{tabular}
}

\newcommand{\ShortCaptionB}{
\begin{tabular}{l|l|llll|llll}
\toprule
 \multicolumn{1}{c}{} & & \multicolumn{4}{c|}{\textbf{Flickr30k}} & \multicolumn{4}{c}{\textbf{COCO}} \\ \cmidrule{3-10} 
 \multicolumn{1}{c}{} & & \multicolumn{2}{c|}{Image-to-Text} & \multicolumn{2}{c|}{Text-to-Image} & \multicolumn{2}{c|}{Image-to-Text} & \multicolumn{2}{c}{Text-to-Image} \\ \cmidrule{3-10} 
 \multicolumn{1}{c}{} & & \multicolumn{1}{c}{R@1} & \multicolumn{1}{c|}{R@5} & \multicolumn{1}{c}{R@1} & \multicolumn{1}{c|}{R@5} & \multicolumn{1}{c}{R@1} & \multicolumn{1}{c|}{R@5} & \multicolumn{1}{c}{R@1} & \multicolumn{1}{c}{R@5} \\ \midrule 
\parbox[t]{2mm}{\multirow{4}{*}{\rotatebox[origin=c]{90}{B/16}}} 
& Pretrained-CLIP\tablefootnote{\href{https://openaipublic.azureedge.net/clip/models/5806e77cd80f8b59890b7e101eabd078d9fb84e6937f9e85e4ecb61988df416f/ViT-B-16.pt}{\nolinkurl{CLIP-ViT-B/16}}} & 0.441 & \multicolumn{1}{l|}{0.682} & 0.248 & 0.451 & 0.517 & \multicolumn{1}{l|}{0.768} & 0.327 & 0.578 \\
& Baseline & 0.470 & \multicolumn{1}{l|}{0.719} & 0.334 & 0.564 & 0.553 & \multicolumn{1}{l|}{0.800} & \textbf{0.409} & \textbf{0.669} \\
& TULIP \cite{tulip} & 0.461 & \multicolumn{1}{l|}{0.708} & \textbf{0.352} & \textbf{0.572} & 0.568 & \multicolumn{1}{l|}{0.803} & 0.407 & 0.661 \\
& \textbf{FineLIP (Ours)} & \textbf{0.528} & \multicolumn{1}{l|}{\textbf{0.753}} & 0.341 & 0.567 & \textbf{0.587} & \multicolumn{1}{l|}{\textbf{0.814}} & 0.404 & 0.662 \\ \midrule
\parbox[t]{2mm}{\multirow{4}{*}{\rotatebox[origin=c]{90}{L/14}}} 
& Pretrained-CLIP\tablefootnote{\href{https://openaipublic.azureedge.net/clip/models/b8cca3fd41ae0c99ba7e8951adf17d267cdb84cd88be6f7c2e0eca1737a03836/ViT-L-14.pt}{\nolinkurl{CLIP-ViT-L/14}}} & 0.485 & \multicolumn{1}{l|}{0.727} & 0.280 & 0.493 & 0.560 & \multicolumn{1}{l|}{0.795} & 0.353 & 0.600 \\
& Baseline & 0.544 & \multicolumn{1}{l|}{0.788} & 0.419 & \textbf{0.653} & 0.608 & \multicolumn{1}{l|}{0.836} & \textbf{0.472} & \textbf{0.721} \\
& TULIP \cite{tulip} & 0.567 & \multicolumn{1}{l|}{0.795} & 0.416 & 0.643 & 0.626 & \multicolumn{1}{l|}{0.847} & 0.461 & 0.711 \\
& \textbf{FineLIP (Ours)} & \textbf{0.622} & \multicolumn{1}{l|}{\textbf{0.828}} & \textbf{0.424} & 0.652 & \textbf{0.634} & \multicolumn{1}{l|}{\textbf{0.848}} & 0.462 & 0.712 \\ \midrule 
\end{tabular}
}

\newcommand{\FullShortCaption}{
\begin{tabular}{l|l|llllll|llllll}
\toprule 
 \multicolumn{1}{c}{} & & \multicolumn{6}{c|}{\textbf{Flickr30k}} & \multicolumn{6}{c}{\textbf{COCO}} \\ \cmidrule{3-14} 
 \multicolumn{1}{c}{} & & \multicolumn{3}{c|}{Image-to-Text} & \multicolumn{3}{c|}{Text-to-Image} & \multicolumn{3}{c|}{Image-to-Text} & \multicolumn{3}{c}{Text-to-Image} \\ \cmidrule{3-14} 
 \multicolumn{1}{c}{} & & \multicolumn{1}{c}{R@1} & \multicolumn{1}{c}{R@5} & \multicolumn{1}{c|}{R@10} & \multicolumn{1}{c}{R@1} & \multicolumn{1}{c}{R@5} & \multicolumn{1}{c|}{R@10} & \multicolumn{1}{c}{R@1} & \multicolumn{1}{c}{R@5} & \multicolumn{1}{c|}{R@10} & \multicolumn{1}{c}{R@1} & \multicolumn{1}{c}{R@5} & \multicolumn{1}{c}{R@10} \\ \midrule 
\parbox[t]{2mm}{\multirow{4}{*}{\rotatebox[origin=c]{90}{B/16}}} 
& Baseline$\ddag$ & 0.0002 & 0.0014 & \multicolumn{1}{l|}{0.0024} & 0.0001 & 0.0007 & 0.0012 & 0.0012 & 0.0082 & \multicolumn{1}{l|}{0.0134} & 0.0011 & 0.0043 & 0.0076 \\
& Baseline & 0.470 & 0.719 & \multicolumn{1}{l|}{0.805} & 0.334 & 0.564 & 0.659 & 0.553 & 0.800 & \multicolumn{1}{l|}{0.869} & 0.409 & 0.669 & 0.766 \\
& SPARC \cite{sparc} & 0.318 & 0.567 & \multicolumn{1}{l|}{0.677} & 0.273 & 0.493 & 0.591 & 0.451 & 0.718 & \multicolumn{1}{l|}{0.817} & 0.366 & 0.629 & 0.732 \\
& LAPS \cite{laps} & 0.541 & 0.765 & \multicolumn{1}{l|}{0.838} & 0.352 & 0.577 & 0.668 & 0.594 & 0.821 & \multicolumn{1}{l|}{0.889} & 0.412 & 0.671 & 0.768 \\
& \textbf{Our Model} & 0.528 & 0.753 & \multicolumn{1}{l|}{0.829} & 0.341 & 0.567 & 0.658 & 0.587 & 0.814 & \multicolumn{1}{l|}{0.882} & 0.404 & 0.662 & 0.760 \\ \midrule 
\parbox[t]{2mm}{\multirow{4}{*}{\rotatebox[origin=c]{90}{L/14}}} 
& Baseline & 0.544 & 0.788 & \multicolumn{1}{l|}{0.862} & 0.419 & 0.653 & 0.739 & 0.608 & 0.836 & \multicolumn{1}{l|}{0.899} & 0.472 & 0.721 & 0.809 \\
& SPARC \cite{sparc} & 0.359 & 0.623 & \multicolumn{1}{l|}{0.730} & 0.289 & 0.513 & 0.610 & 0.500 & 0.773 & \multicolumn{1}{l|}{0.859} & 0.383 & 0.648 & 0.749 \\
& LAPS \cite{laps} & 0.624 & 0.834 & \multicolumn{1}{l|}{0.894} & 0.422 & 0.649 & 0.733 & 0.643 & 0.856 & \multicolumn{1}{l|}{0.915} & 0.462 & 0.716 & 0.803 \\
& \textbf{Our Model} & 0.622 & 0.828 & \multicolumn{1}{l|}{0.888} & 0.424 & 0.652 & 0.736 & 0.634 & 0.848 & \multicolumn{1}{l|}{0.910} & 0.462 & 0.712 & 0.801 \\ \midrule 
\parbox[t]{2mm}{\multirow{3}{*}{\rotatebox[origin=c]{90}{bigG/14}}} 
& Baseline & 0.536 & 0.782 & \multicolumn{1}{l|}{0.859} & 0.443 & 0.673 & 0.755 & 0.608 & 0.838 & \multicolumn{1}{l|}{0.905} & 0.505 & 0.753 & 0.833 \\
& LAPS \cite{laps} & 0. & 0. & \multicolumn{1}{l|}{0.} & 0. & 0. & 0. & 0. & 0. & \multicolumn{1}{l|}{0.} & 0. & 0. & 0. \\
& \textbf{Our Model} & 0.541 & 0.776 & \multicolumn{1}{l|}{0.850} & 0.394 & 0.620 & 0.706 & 0.637 & 0.852 & \multicolumn{1}{l|}{0.911} & 0.468 & 0.722 & 0.810 \\ \bottomrule 
\end{tabular}
}

\newcommand{\CLIMAbalation}{
\begin{tabular}{c|llll|llll}
\toprule
 & \multicolumn{4}{c|}{\textbf{Urban1K}} & \multicolumn{4}{c}{\textbf{DOCCI}} \\ \cmidrule{2-9}
\multicolumn{1}{c|}{} & \multicolumn{2}{c|}{Image-to-Text} & \multicolumn{2}{c|}{Text-to-Image} & \multicolumn{2}{c|}{Image-to-Text} & \multicolumn{2}{c}{Text-to-Image} \\ \cmidrule{2-9} 
 & \multicolumn{1}{c}{R@1} & \multicolumn{1}{c|}{R@5} & \multicolumn{1}{c}{R@1} & \multicolumn{1}{c|}{R@5} & \multicolumn{1}{c}{R@1} & \multicolumn{1}{c|}{R@5} & \multicolumn{1}{c}{R@1} & \multicolumn{1}{c}{R@5} \\ \midrule
ATRM + FILIP & 0.315 & \multicolumn{1}{l|}{0.525} & 0.436 & 0.657 & 0.354 & \multicolumn{1}{l|}{0.620} & 0.426 & 0.694 \\ 
ATRM + CLIM (Ours) & \textbf{0.907} & \multicolumn{1}{l|}{\textbf{0.983}} & \textbf{0.893} & \textbf{0.975} & \textbf{0.771} & \multicolumn{1}{l|}{\textbf{0.954}} & \textbf{0.795} & \textbf{0.958} \\ \bottomrule 
\end{tabular}
}
\begin{abstract}
As a pioneering vision-language model, CLIP (Contrastive Language-Image Pre-training) has achieved significant success across various domains and a wide range of downstream vision-language tasks. However, the text encoders in popular CLIP models are limited to processing only 77 text tokens, which constrains their ability to effectively handle longer, detail-rich captions. Additionally, CLIP models often struggle to effectively capture detailed visual and textual information, which hampers their performance on tasks that require fine-grained analysis. To address these limitations, we present a novel approach, \textbf{FineLIP}, that extends the capabilities of  CLIP. FineLIP enhances cross-modal text-image mapping by incorporating \textbf{Fine}-grained alignment with \textbf{L}onger text input within the CL\textbf{IP}-style framework. FineLIP first extends the positional embeddings to handle longer text, followed by the dynamic aggregation of local image and text tokens. The aggregated results are then used to enforce fine-grained token-to-token cross-modal alignment. We validate our model on datasets with long, detailed captions across two tasks: zero-shot cross-modal retrieval and text-to-image generation. Quantitative and qualitative experimental results demonstrate the effectiveness of FineLIP, outperforming existing state-of-the-art approaches. Furthermore, comprehensive ablation studies validate the benefits of key design elements within FineLIP. The code will be available at  \url{https://github.com/tiiuae/FineLIP}.

\end{abstract}

\section{Introduction}
\label{sec:intro}

\begin{figure}[t]
  \centering
   \includegraphics[width=\columnwidth]{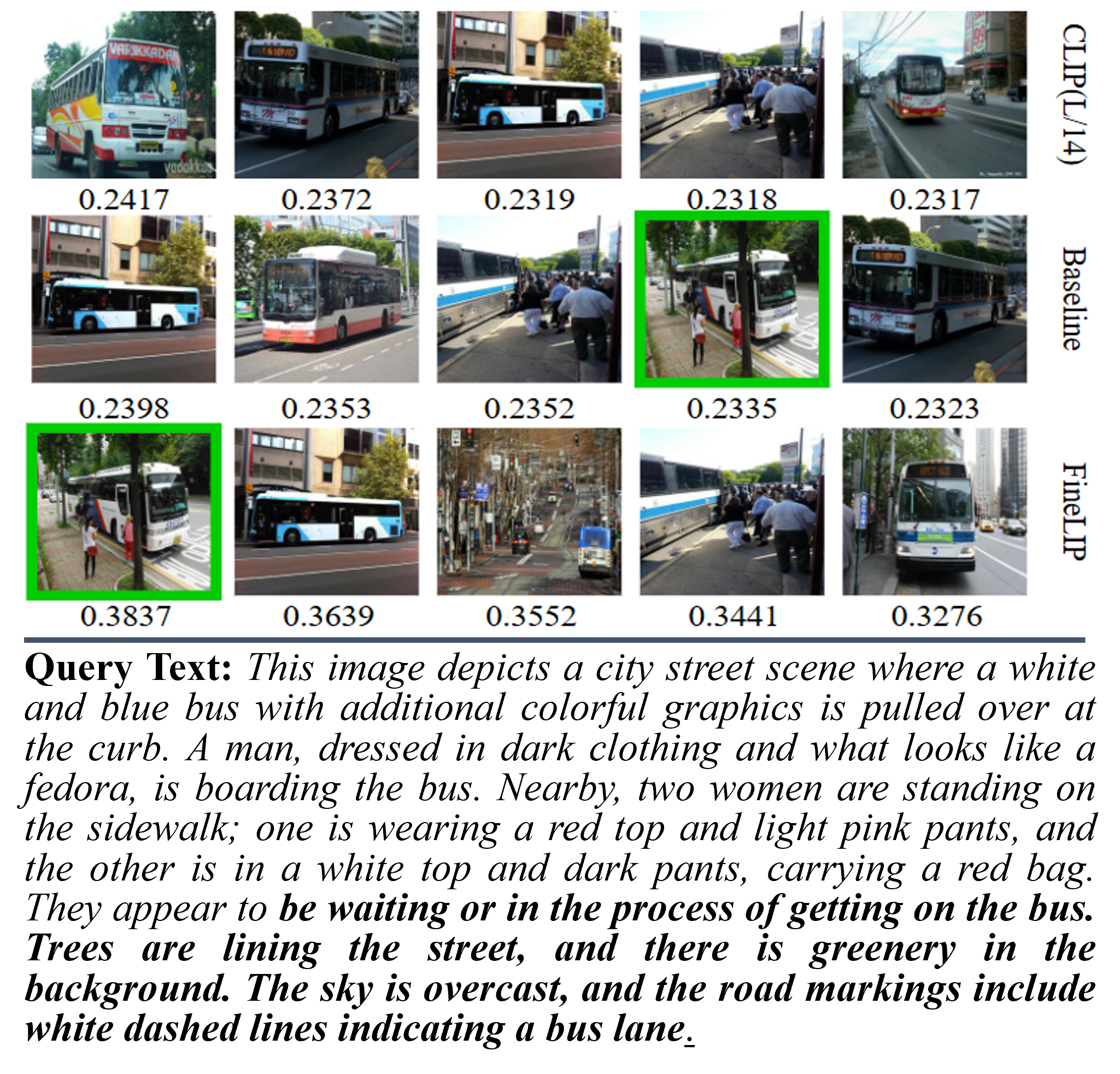}
   \caption{Top-5 text-to-image retrieval results on \textit{Urban1k} dataset \cite{longclip} for L/14 variants of CLIP, Baseline and FineLIP (Ours), with image retrieval scores. The correct retrieved images are marked with green boxes. CLIP ignores the caption in bold due to the 77-token limit.}
   \label{fig:intro}
\end{figure}

Contrastive Language-Image Pre-training (CLIP) \cite{clip} has set a high standard for vision-language models (VLMs), particularly by leveraging a dual-stream architecture with separate encoders for visual and textual inputs. Trained on 400 million web-sourced image-text pairs, CLIP aligns positive pairs in the embedding space while pushing negative pairs apart through contrastive loss, contributing to its strong zero-shot performance across various downstream tasks. Furthermore, CLIP’s vision encoder has become a foundational component in many large vision-language models (LVLMs) \cite{llava,sharegpt}, serving as a key tool for extracting visual features. While CLIP has been successful in many areas, it is primarily trained with image-caption pairs where the captions are typically short. However, recent studies reveal that images often convey richer visual details than the short captions can encapsulate \cite{dci,dreamlip,longclip}, including intricate relationships and attributes like colors, locations, and sizes \cite{docci}. In response, new datasets have been introduced, pairing images with longer and more detailed captions, created either through manual annotation or by leveraging LVLMs' captioning capabilities\cite{dci,docci,iiw,sharegpt}. This shift underscores the need to explore methods for integrating extended captions into CLIP-style models. 

To extend CLIP models for handling long and descriptive captions, there are significant limitations. The primary challenge arises from CLIP's 77-token processing limit. Even within this limit, positional embeddings for higher token indices remain under-trained \cite{longclip} due to the prevalence of shorter captions in CLIP's training data, further diminishing its ability to fully leverage detailed captions. In text-to-image retrieval, for example, many words in the long caption cannot be encoded, leading to poor retrieval performance, as shown in the top row of \cref{fig:intro}. Extending CLIP’s token limit could enable training with longer captions and capture more detailed descriptions. However, increasing the token capacity would disrupt the existing pre-trained CLIP weights alignment, as they are optimized for the 77-token limit. This change would require starting from scratch, forfeiting the benefits of CLIP's extensive pre-training on 400 million image-text pairs. To address this limitation, we propose an approach that leverages longer captions for training while preserving the strengths of the original pre-trained model, thereby improving performance without sacrificing the learned visual-textual alignment, as demonstrated in the second row of \cref{fig:intro}. 

The second shortcoming of CLIP when handling long captions stems from its reliance on loss design, which primarily focuses on aligning global visual and text features. This approach limits the model’s ability to capture fine-grained details that are essential for understanding richer, more complex captions. Notably, longer, detailed captions provide richer contextual and fine-grained information, while images contain local details beyond what global features can capture. Despite these advantages, most methods that incorporate long captions into training, such as \cite{longclip, tulip}, typically rely on global features alone. For instance, Long-CLIP \cite{longclip} extracts global features for both long and short captions and aligns them with visual features and their PCA components using contrastive losses. Similarly, TULIP \cite{tulip} aligns a single global visual feature with both caption types by applying a balancing weight. These methods focus exclusively on global features, neglecting local details that are crucial for capturing finer image-text relationships. In contrast, we propose leveraging image and text local features through fine-grained, token-level cross-modal alignment, enabling a more effective capture of these rich details. As illustrated in \cref{fig:intro}, our approach, which integrates fine-grained alignment, surpasses the baseline (global features only) in top-5 text-to-image retrieval, demonstrating the advantage of token-level alignment in enhancing detailed retrieval performance.

In this paper, we introduce FineLIP, a novel approach that achieves \textbf{Fine}-grained alignment with \textbf{L}onger text input within the CL\textbf{IP} framework, enabling tasks that require detailed multi-modal comprehension. FineLIP extends CLIP’s capabilities by surpassing the 77-token limit, allowing for longer, more descriptive captions to be used. Additionally, FineLIP incorporates a new token-to-token alignment strategy, fully leveraging local visual and text features. This enhancement significantly improves fine-grained alignment between visual and textual modalities, enabling more effective extraction of nuanced details often overlooked by existing methods. Our main contributions include: (1) Introducing a token refinement and alignment mechanism that facilitates fine-grained, token-level contrastive learning between visual and textual elements, effectively addressing CLIP’s token limitation and enhancing cross-modal alignment for longer captions. (2) Conducting rigorous evaluations of FineLIP on extensive datasets, demonstrating significant performance improvements over recent SOTA methods. (3) Providing comprehensive ablation studies to validate the contribution of each component in FineLIP, highlighting the roles of key elements. 
\section{Related Works}
\label{sec:related_works}

\begin{figure*}[t]
    \centering
    \includegraphics[width=\textwidth]{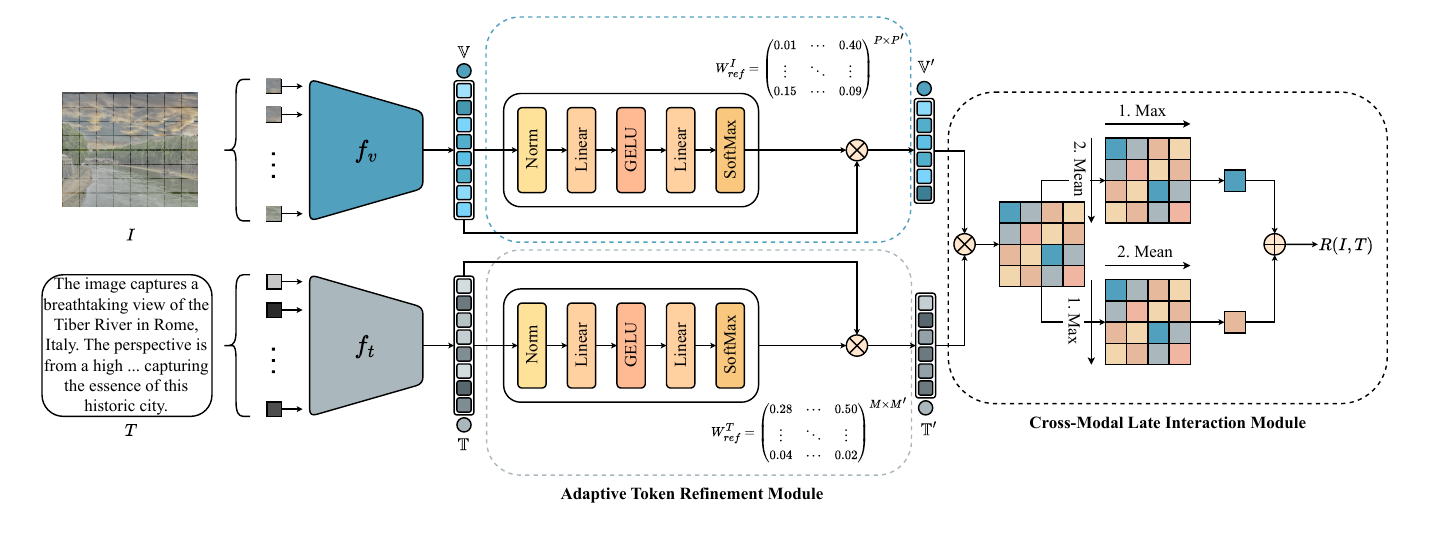}
    \vspace{-20pt} 
    \caption{Overview of \textbf{FineLIP} architecture. Image-caption pair $(I,T)$ are passed through their respective encoders $f_v$ and $f_t$ to obtain the embeddings $\mathbb{V}$ and $\mathbb{T}$. The embeddings then are fed to the Adaptive Token Refinement Module to dynamically aggregate them into a set of representations $\mathbb{V'}$ and $\mathbb{T'}$ that offer improved information density. Finally, these aggregated tokens are forwarded into the Cross-Modal Late Interaction Module to achieve token-to-token fine-grained alignment. Note that the full caption $T$ is shortened for display.}
    \label{fig:method}
\end{figure*}

{\bf Vision-Language Foundational Models:} In recent years, VLMs have garnered significant interest and made remarkable progress. A prominent area of VLM research focuses on contrastive learning-based models \cite{align,clip,filip}, which adopt a contrastive loss to optimize image/text representations. The goal is to maximize the similarity of matched image-caption pairs while minimizing the similarity of mismatched pairs. CLIP \cite{clip} stands out as a pioneering work, demonstrating impressive performance on downstream tasks, such as zero-shot classification and image-text retrieval. Additionally, CLIP serves as the visual encoder in many mainstream multi-modal foundation models \cite{llava, sharegpt}. OpenAI has released a series of CLIP checkpoints, trained on a large dataset of 400 million image-text (primarily short) pairs \cite{openai2021clip}. Following CLIP, numerous works have been proposed to improve it from various perspectives. In this paper, we focus on the extension of CLIP to handle longer captions.

{\bf CLIP with longer captions:} With the progress in LVLMs \cite{llava,sharegpt}, their image captioning capabilities have been impressively enhanced, enabling the generation of longer and more detailed captions from images. Some efforts also focus on manually annotating images with high-quality, detailed captions \cite{docci, dci}. To effectively leverage these long and detailed captions, Long-CLIP \cite{longclip} builds upon CLIP by extending the text token length from 77 to 248. Furthermore, global image features and their PCA decomposed features are aligned with detailed and summarized short captions, respectively. TULIP \cite{tulip} also extends the context window by incorporating relative positional encodings into CLIP. Dreamlip \cite{dreamlip} splits long captions into short individual captions. These short captions are then considered as multiple positive text samples, and the model optimization employs a multi-positive contrastive learning approach. However, all evaluations in Dreamlip focus on tasks involving short captions. Its performance on datasets with long captions remains unclear. A recent study \cite{dci} highlights the importance of dense captions and introduces a manually labeled dataset. However, they use a large language model (LLM) \cite{llama} to summarize the long captions into shorter versions before inputting them into CLIP for fine-tuning, leaving the direct application of dense, long captions largely unexplored. In our work, we focus primarily on the direct use of long captions and evaluate our model on tasks involving long captions.


{\bf Fine-grained understanding in vision-language models:} In many CLIP variants, cross-modal interaction is confined to a coarse level of granularity, as they primarily align global image and text representations within a shared latent space. Although this approach is effective for specific tasks, these models often struggle with tasks that require fine-grained features, such as localization \cite{perceptual}, and understanding spatial relationships \cite{valse}. To address the overlooked fine-grained visual and textual information in existing CLIP models, recent research has emerged that emphasizes the alignment between local image and text tokens, enabling the learning of representations with finer details \cite{filip, pyramidclip, sparc, laps}. FILIP \cite{filip} introduces a cross-modal interaction mechanism that optimizes the token-wise similarity between visual and textual tokens. SPARC \cite{sparc} groups image patches corresponding to individual language tokens using a sparse similarity metric. This method contrasts language-grouped vision embeddings with text token embeddings. Similarly, Fu \etal \cite{laps} propose selecting visual patches driven by language supervision before applying cross-modal token-to-token contrastive learning. However, these methods are designed and validated specifically for short captions. Moreover, they focus on refining only the visual representations, leaving the raw textual representations unchanged. In contrast, our approach not only enforces fine-grained alignment for longer captions but also addresses the ambiguity often present in raw text representations. 
\section{Methodology} \label{sec:methodology}
In FineLIP, we focus on enhancing CLIP-style models for longer captions. First, we extend the positional embeddings and initialize our model using a pretrained CLIP model. Next, we aggregate image and text tokens separately to obtain informative representations for each modality. Finally, we introduce fine-grained, token-level cross-modal matching to learn aligned visual and textual features. This enhances downstream tasks like image-text retrieval and text-to-image generation. The flowchart is presented in \cref{fig:method}, 

{\bf Preliminary:}
For an image-text pair $(I,T)$, let $I$ represent the image input and $T$ denote the corresponding text input. 
Let $f_v(\cdot)$ be the CLIP image encoder (Vision Transformer version \cite{vit}) and $f_t(\cdot)$ be the CLIP text encoder. 
The image $I$ is partitioned into $P$ non-overlapping patches, each representing a distinct region of the visual input. 
These patches are processed through $f_v(\cdot)$ to yield a set of visual embeddings $\mathbb{V}=\{v_{cls},v_1,\ldots,v_P\} \in \mathbb{R}^{(P+1)\times d}$. 
Here, $v_{cls}$ denotes the [CLS] token, which encapsulates global information, $v_i$ represents the patch embedding containing the local information of the image, and $d$ denotes the feature dimension.
Similarly, the text input $T$ is tokenized, padded, and fed into $f_t(\cdot)$, producing a set of textual features $\mathbb{T}=\{t_0,t_1,\ldots,t_{eos},\ldots,t_{247}\} \in \mathbb{R}^{248\times d}$. 
Here, $t_0$ and $t_{eos}$ represent the features for the [BOS] and [EOS] tokens, respectively.
We set a maximum token limit of 248 to accommodate long text lengths. 
Tokens between $t_0$ and $t_{eos}$ are considered valid input tokens, while tokens beyond $t_{eos}$ up to $t_{247}$ are padding.

\subsection{Stretching of Positional Embeddings}
\label{subsec:stretching_ps}
In CLIP’s original architecture, the input text length is capped at 77 tokens because of its learned absolute positional embeddings. 
This constraint limits the model's ability to process longer and more detailed textual descriptions. 
As shown in the ablation study in \cref{tab:ablation1}A, the preferred approach is to encode longer text while leveraging the capabilities learned in the pretrained CLIP. 
In this work, we address this limitation by employing knowledge-preserved stretching of the positional embeddings \cite{longclip}. 
Following \cite{longclip}, we preserve the positional embeddings of the first 20 tokens, which have been empirically identified as well-trained during CLIP's pretraining. 
For token positions beyond 20, we apply an adaptive interpolation method to stretch the positional embeddings to $4 \times$ their original length. 
As a result, the stretched embeddings reach a length of $20 + (77 - 20) \times 4 = 248$, which is typically sufficient to encode longer captions. 
This adjustment allows longer text inputs while minimizing disruptions to the strong cross-modal alignment achieved in the pretrained CLIP.

\subsection{Adaptive Token Refinement Module}
In vision-language tasks, individual tokens often lack meaningful information on their own and require context from surrounding tokens for complete comprehension. 
In the CLIP framework, where the image and text encoders are transformer-based, the representation of each local token in the final layer can remain ambiguous. 
Before applying token-to-token fine-grained alignment, refining the tokens to reduce this ambiguity is essential for enabling effective cross-modal correspondence. 

Within the domain of token refinement, some approaches focus on selecting tokens \cite{adaViT,Dynamicvit,laps,tsm,tome}. 
These methods leverage the observation that token similarity increases in deeper layers, indicating that only a subset of tokens carries decision-relevant information. 
While token selection can reduce redundancy and enhance training and inference efficiency, especially when applied from early layers, we argue that this approach may still result in some information loss, even when using comprehensive metrics for selection. 
Thus, our work, FineLIP, adopts token aggregation over selection to minimize ambiguity while avoiding information loss. A distinguishing feature of our aggregation approach is that refinement is applied to both image and text tokens, rather than focusing solely on the visual branch. 
This is motivated by the observation that ambiguity exists in both visual and text tokens, albeit to a lesser extent in the latter.


After extracting visual ($\mathbb{V}$) and textual ($\mathbb{T}$) features through their respective encoders, we introduce the Adaptive Token Refinement Module (ATRM) for further processing. 
This module dynamically refines tokens by aggregating the original token sets to produce more informative and meaningful representations. 
In contrast to heuristic approaches that merge tokens based on predefined rules \cite{tsm,sparc,tome}, the aggregation in ATRM is learned automatically. 
The refinement process starts with an input of $N$ tokens, each of dimension $d$, represented as $X = \{x_1, x_2, \dots, x_N\} \in \mathbb{R}^{N \times d}$.
ATRM dynamically aggregates tokens to generate a condensed set of $N'$ refined tokens, denoted as $X' = \{x'_1, x'_2, \dots, x'_{N'}\}$, where $N' < N$. 
The ratio $\frac{N'}{N}$ defines the aggregation ratio (set as 0.2 by default), controlling the extent of tokens compression. 
This transformation is expressed as $X' = W_{ref}\cdot X$, where $W_{ref} \in \mathbb{R}^{N' \times N}$ is a learnable matrix, with the condition $\sum_{i=1}^{N'} (W_{ref})_{i,j} = 1$.  
This design ensures the refinement process is fully differentiable, enabling seamless integration into end-to-end learning frameworks. ATRM computes $W_{ref}$ in a manner inspired by self-attention yet optimized for efficiency. 
Specifically, $W_{ref}=\textit{SoftMax}(\frac{W_q\sigma(XW_k)^T}{\tau})$, where $W_k \in \mathbb{R}^{d \times d_k}$ and $W_q \in \mathbb{R}^{N' \times d_k}$ are trainable projection matrices ($d_k < d$), $\sigma$ is a nonlinear function (GELU \cite{gelu}) and $\tau$ is a learnable temperature parameter that encourages sparse attention. The "Adaptive Token Refinement Module" block in \cref{fig:method} illustrates how the aggregation is implemented with neural network layers. Note that aggregation is applied to both the visual and textual branches, where the ATRM modules share the same architecture but use separate parameters for each branch.

By aggregating tokens into a more discriminative and informative set, ATRM effectively reduces the number of tokens while ensuring that each token in the final set retains key semantic content. This lays the foundation for later fine-grained cross-modal alignment.
As a result of the refinement process , the visual features $\mathbb{V}$ are updated to $\mathbb{V'} = \{v_{cls},v'_1, \ldots, v'_{P'}\}$ and the textual features $\mathbb{T}$ are updated to $\mathbb{T'} = \{t'_1, \ldots, t'_{M'},t_{eos}\}$. Here ${P'}$ and ${M'}$ are the numbers of aggregated visual and textual tokens, respectively.
The $v_{cls}$ token from the visual input and the $t_{eos}$ token from the text input, which provides global representations, are not used in the refinement but are retained in the updated sets ($\mathbb{V'}$ and $\mathbb{T'}$). Additionally, any padded tokens in $\mathbb{T}$ are ignored and do not participate in token refinement.
With the ATRM, each refined token possesses a higher information density, easing the token-to-token alignment introduced later. 
Additionally, reducing the number of tokens contributes to improved computational efficiency. 

\subsection{Fine-Grained Cross-Modal Alignment}
Fine-grained alignment is crucial in vision-language tasks due to the nuanced relationships between visual elements and their textual descriptors, such as spatial and semantic relationships.  
Traditional coarse-grained alignment methods focus on optimizing the similarity of global features from text and vision, which can result in lost details \cite{filip}.
In contrast, fine-grained alignment allows the model to focus on the intricate interactions between local visual tokens and their corresponding textual tokens. 
This integration of fine-grained alignment enhances the model's ability to discern subtle cross-modal relationships, improving its performance on complex tasks. 

In FineLIP, we employ the Cross-Modal Late Interaction module (CLIM) after ATRM to achieve a more precise alignment that reflects the detailed correspondence between image and text embeddings. 
With the refined set of visual tokens $\mathbb{V'}$ and textual tokens $\mathbb{T'}$ from ATRM, we first compute the cosine similarity $S(v'_i, t'_j)$ between each visual token $v'_i$ and each text token $t'_j$.
This token-level similarity captures the detailed correspondence between specific parts of the image and individual text tokens, allowing for alignment at a finer granularity. 
We then use the following pooling strategies to obtain the overall alignment score,  
\begin{equation}
    R(I,T) = \frac{1}{P'} \sum_{i=1}^{P'} \max_j S(v'_i, t'_j) 
    + \frac{1}{M'} \sum_{j=1}^{M'} \max_i S(t'_i, v'_j)
\label{eq:relavance_score}
\end{equation}
where the first and the second terms represent the image-to-text and text-to-image fine-grained similarity scores, respectively. 
This solution captures bidirectional alignments, ensuring that image and text are accurately represented in relation to each other.

Instead of the traditional contrastive loss \cite{clip}, we employ Triplet Marginal Loss \cite{triplet} for both image-to-text and text-to-image queries. 
The triplet loss is advantageous because it ensures that the similarity scores of positive pairs exceed those for negative ones by a predefined margin, allowing closer proximity of features from positive pairs and greater separations of features from negative pairs.
For image-to-text (text-to-image) alignment, given a query image (text) $I_q$, a positive text (image) $T^+$, and a negative text (image) $T^-$, the loss is formulated as
\begin{equation}
\begin{split}   
    &\mathcal{L}_{i2t} = \max (0, R(I_q,T^-) - R(I_q,T^+) + \alpha) \\
    &\mathcal{L}_{t2i} = \max (0, R(T_q,I^-) - R(T_q,I^+) + \alpha) \\
    &\mathcal{L}_{triplet} = \mathcal{L}_{i2t} + \mathcal{L}_{t2i}
\end{split}
\label{eq:loss}
\end{equation}
where $\alpha$ is the margin that separates positive and negative pairs. We choose $\alpha$ as 0.2.
This dual-triplet loss encourages effective optimization of both image-to-text and text-to-image alignments, prompting the model to accurately rank positive pairs (\ie images with their corresponding text and vice versa) over negative pairs. 
Combining bidirectional similarity scores and optimizing through this comprehensive triplet loss, our model achieves enhanced fine-grained and precise token-level cross-modal alignment in both directions.
Furthermore, it is essential to note that the tokens capturing global features (\ie [CLS] token for image and [EOS] token for text) are also involved in the loss, as defined in  $\mathbb{V'}$ and $\mathbb{T'}$. 
This design preserves features of different granularity within the same modality and facilitates cross-granular cross-modal alignment, whose effectiveness will be validated in the next section. 


\section{Experiments}\label{sec:experiments}
\begin{table*}[t]
    \centering
    \resizebox{2\columnwidth}{!}{\LongCaption}
    \caption{Comparison of zero-shot cross-modal retrieval on long caption datasets. \textbf{FineLIP*} shares the same trained model as  \textbf{FineLIP}, but just differs in the inference. In \textbf{FineLIP*}, we compute cross-modal similarities using coarse-grained and fine-grained features separately (since both global and local features are used in the training, see \cref{sec:methodology}), and then subsequently combine them through a weighted sum. \textbf{FineLIP*} improves the overall robustness and adaptability in various retrieval tasks. Best is shown in bold}
    \label{tab:long_caption}
\end{table*}

\subsection{Experimental Setting}
{\bf Training Dataset:} Following \cite{longclip}, we use the \textit{ShareGPT4V} dataset \cite{sharegpt} for training, which contains approximately 1.2 million image-caption pairs. 
This dataset is particularly well-suited for long caption tasks, due to its rich, detailed captions that capture object properties, spatial relationships, and other nuanced scene-specific information.
The average caption length in \textit{ShareGPT4V} is approximately 143.6 words, which is significantly longer compared to captions in commonly used datasets like \textit{CC12M} (20.2 words) and \textit{COCO} (10.2 words). 

\noindent{\bf Evaluation Tasks, Datasets, and Metrics:} To align with our focus on longer captions, we evaluate our model on two tasks: zero-shot long caption cross-modal retrieval and long-text-to-image generation.
\begin{itemize}
    \item {\bf Zero-shot Long Caption Cross-Modal Retrieval:} We use the \textit{Urban1k} \cite{longclip} and \textit{DOCCI} \cite{docci}, datasets, which contain long, detailed captions. 
    Full dataset details are available in the supplementary material.
    We report Recall@1, Recall@5, and Recall@10 metrics for image-to-text (I2T) and text-to-image (T2I) retrieval.
    \item {\bf Long Text-To-Image Generation:} This task also utilizes \textit{Urban1k} and \textit{DOCCI}. 
    For generation, we leverage the Stable Diffusion XL (SDXL) \cite{SDXL} model, which features three significant enhancements over its predecessors: a substantially larger UNet architecture, dual text encoders that expand parameter capacity, and a two-stage refinement process that adds high-quality details. 
    This robust framework positions SDXL as an ideal choice for generating images. 
    To objectively assess image generation quality, we employ the Fréchet Inception Distance (FID) \cite{fid} metric.
\end{itemize}

\noindent{\bf Implementation details}
Our model is initialized with CLIP's pre-trained weights and then fine-tuned on the \textit{ShareGPT4V} dataset for $6$ epochs using \textit{Nvidia A10} GPUs, with a global batch size of $128$ ($64$ for bigG/14).
We employ two learning rates: $1\text{e-}6$ for the CLIP encoders and $2\text{e-}4$ for our ATRM.
We experiment with three different CLIP architecture, specifically \textit{CLIP-ViT-B/16} (B/16), \textit{CLIP-ViT-L/14} (L/14), and \textit{CLIP-ViT-bigG/14} (bigG/14), to comprehensively assess performance across various model capacities. 
In our evaluation, we compare our approach with five methods: Baseline, SPARC \cite{sparc}, LAPS \cite{laps}, Long-CLIP \cite{longclip}, and TULIP \cite{tulip}. 
The Baseline refers to the configuration in which we apply the same positional embedding stretching as in \textbf{FineLIP} while utilizing traditional cross-modal coarse-level contrastive learning for training.
\cite{longclip} and \cite{tulip} are the typical works that utilize long captions directly in both training and evaluation. Although SPARC and LAPS are primarily designed for training and evaluating with short captions, they are representative works for cross-modal fine-grained alignment. Thus, we re-implement these algorithms and employ the same training and evaluation protocols used in FineLIP. The results for \cite{longclip} and \cite{tulip} are taken directly from their original papers, as they use the same dataset for training as FineLIP.

\subsection{Results}
\noindent{\bf Zero-shot Long Caption Cross-Modal Retrieval:} \cref{tab:long_caption} shows the image-text retrieval scores on long caption datasets \textit{Urban1k} and \textit{DOCCI}. 
Our model achieves the best performance across all metrics in image-to-text (I2T) and text-to-image (T2I) retrievals. Specifically, when using B/16 as the backbone, our model achieves an I2T performance of 0.907 for R@1 on the \textit{Urban1k} dataset, which represents nearly a 5\% improvement over the Baseline model's score of 0.859. 
\begin{table}[t]
    \centering
    \fid
    \caption{FID scores for various models (L/14 variant) utilizing the text encoder in the Stable Diffusion XL (SDXL) pipeline for image generation. The scores reflect the quality of generated images, with lower FID values indicating better performance.}
    \label{tab:fid}
\end{table}
A similar trend is observed on the more challenging \textit{DOCCI} dataset (0.771 \vs 0.731). 
Regarding T2I retrieval, \cref{tab:long_caption} also displays considerable gains on both datasets, although the improvement is somewhat less pronounced compared to I2T retrieval.
The comparison with baseline, where only coarse-level features are aligned, underscores the necessity of fine-grained alignments to capture the complex associations between images and their corresponding detailed textual descriptions. Long-CLIP yields poorer retrieval results than the Baseline, while TULIP shows performance on par with the Baseline. However, both consistently lag behind FineLIP across all metrics.
Compared to other state-of-the-art approaches like SPARC and LAPS, which incorporate fine-grained alignment in different ways, our model shows non-trivial improvements, thanks to the innovative design of the proposed ATRM and CLIM components and their combination. 
Similar improvements are observed when switching the backbone to L/14 or bigG/14, demonstrating the effectiveness of FineLIP, regardless of model size. The supplementary material provides visualization of different models for both I2T and T2I tasks. Additionally, we present evaluation results on zero-shot short-caption image-text retrieval across various models in the supplementary material, offering further insights into the significance of long captions.

\begin{figure*}[t]
  \centering
  \resizebox{\textwidth}{7.8cm}{
   \includegraphics{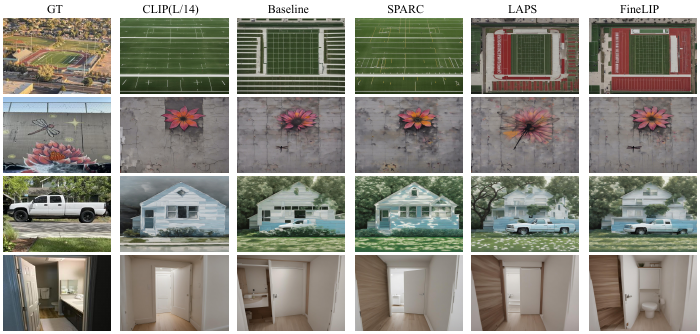}}
   \caption{Visualization of long text-to-image generations using different L/14 variants. GT means the ground-truth images paired with the captions in image generation. Zoom in for better visualization. Note that the captions used as well as the detailed analysis of these examples are included in the supplementary material.}
   \label{fig:generation}
\end{figure*}

\noindent{\bf Long-Text-to-Image Generation:} Our model integrates seamlessly into the SDXL pipeline, enhancing its capability for long-text image generation tasks.
Stable Diffusion models traditionally leverage the \textit{CLIP-ViT-L/14} text encoder to extract features from text inputs, which guide the image generation process. 
However, this text encoder is limited to processing only 77 tokens, constraining its ability to handle longer descriptions. Following \cite{longclip, tulip}, we replace the original CLIP text encoder in SDXL with the FineLIP model’s text encoder.
This allows SDXL to process extended text inputs that exceed the token limit, enabling the generation of high-fidelity images that accurately capture the complexities and fine-grained details found in longer captions. The performance metrics for various models are summarized in \cref{tab:fid}, showcasing the effectiveness of our approach, especially when compared to the Baseline. \cref{fig:generation}, which displays typical images generated by various models, demonstrates that FineLIP effectively generates high-quality images that capture global and fine-grained details from extended text inputs. The specific texts used and detailed analysis can be found in the supplementary material.

\subsection{Ablation Study}
We conduct extensive ablation studies to verify the effectiveness of different components in FineLIP, all evaluated on the zero-shot image-text retrieval tasks. The B/16 architecture is used in all experiments.

\noindent{\bf The impact of Initialization and Positional Embedding Stretching.}
As highlighted in \cref{sec:intro}, a straightforward approach to training a CLIP-style model for long captions is to initialize the model randomly and set a higher token limit. However, our empirical results in \cref{tab:ablation1}A indicate that this method, labeled as Baseline\_rand\_init, results in poor performance. In contrast, the Pretrained-CLIP model, despite its 77-token limitation and being trained exclusively on short captions, achieves significantly better results on tasks involving long captions (with text exceeding the 77-token limit truncated in this case). Thereafter, we are motivated to leverage the pretrained CLIP weights and adapt them for handling long captions. Baseline\_nPE is the approach where we finetune the Pretrained-CLIP model directly on long captions dataset. Following \cite{lit}, we also tried training the long-context text encoder (by setting a higher token limit) from scratch while reusing the weights of CLIP’s vision encoder (Baseline\_Lv). Baseline in \cref{tab:ablation1}A, as outlined in \cref{subsec:stretching_ps}, we extend the token limit from 77 to 248 by positional embedding stretching, while initializing the model with Pretrained-CLIP weights. Clearly, Baseline constantly achieves the best performances across all cases. This further confirms the importance of stretching the positional embeddings and fully utilizing pretrained CLIP for enhanced results on long caption tasks.

\noindent{\bf The impact of ATRM.} 
We conduct studies across multiple configurations to evaluate the impact of our adaptive token refinement module (ATRM) before implementing the token-level fine-grained cross-modal alignment, with results presented in \cref{tab:ablation1}B.
\begin{table}[t]
    \centering
    \resizebox{\columnwidth}{!}{\ablation}
    \caption{\textbf{A:} The impact of Initialization and Positional Embedding Stretching. \textbf{B:} The impact of ATRM and CLIM within our model. \textbf{C:} Effect of different aggregation ratios in ATRM. \textbf{D:} The importance of retaining global features in alignment. ``No Global Tokens" means model trained with $v_{cls}$ and $t_{eos}$ token removed from $\mathbb{V'}$ and $\mathbb{T'}$ during fine-grained cross-modal alignment. Best result is in bold. Zoom in for better visualization.}
    \label{tab:ablation1}
\end{table}

\begin{itemize}
    \item {\bf CLIM Only:} we apply the Cross-Modal Late Interaction Module (CLIM) without any token refinement. 
    \item {\bf ATRM (Image) + CLIM:} In this setup, only image tokens are refined with the ATRM module, while textual tokens remain unchanged. 
    Compared to CLIM-only, this shows substantial gains in T2I retrieval (R@1 improved from 0.782 to 0.890 in \textit{Urban1k}). 
    These improvements confirm that raw vision tokens are ambiguous, and ATRM effectively reduces this ambiguity by aggregating tokens into a more informative set. 
    Consequently, the learned representations become more discriminative, yielding significant gains in cross-modal retrieval. 
    
    \item {\bf ATRM (Text) + CLIM:} Similarly, this setup only refines text tokens before applying CLIM. 
    While noticeable improvements are compared to the CLIM-only baseline, the gains are slightly more moderate than those in the image-only ATRM setup.
    One possible reason is that text tokens (\ie words, sub-word units, \etc) are inherently structured for human communication and generally have higher information density than image tokens \cite{moco}, resulting in lower ambiguity. 
    Therefore, their aggregation is less critical than that of image tokens.
    
    
    \item {\bf ATRM (Both) + CLIM:} Applying ATRM to both image and text tokens yields the best performance across all metrics on both datasets, consistent with our statement in \cref{sec:methodology}. 
    This demonstrates that refining tokens in both modalities provides highly condensed cross-modal tokens, allowing the model to achieve the most effective fine-grained alignment, with I2T R@1 reaching 0.907 on \textit{Urban1k} and T2I R@1 achieving 0.795 on \textit{DOCCI}.  
\end{itemize}


\noindent{\bf Impact of Aggregation Ratio in ATRM.} 
The aggregation ratio in ATRM regulates the level of token refinement, influencing the information density of each refined token. 
We empirically analyze its effect, with findings presented in \cref{tab:ablation1}C. 
Varying the ratio from 0.1 to 0.4, we observe that a ratio of 0.2 yields the best performance across most metrics. 
Increasing the ratio retains more tokens, which reduces the information density of each token. When set to 0.4, performance declines across most metrics. 
Conversely, lowering the aggregation ratio to 0.1 also results in decreased performance, particularly in T2I retrieval. This suggests that an overly aggressive reduction in tokens can limit the model’s ability to retain critical details for effective cross-modal alignment. 
Overall, these findings highlight the importance of the aggregation ratio in ATRM, showing that a balanced value, such as 0.2, optimally removes ambiguity in tokens while preserving essential details, leading to enhanced performances on retrieval tasks with long captions.

\noindent{\bf Effectiveness of retaining global features in loss}. 
As mentioned in \cref{sec:methodology}, global features (\ie [CLS] token for image and [EOS] token for text) are also retained in the alignment score calculation \cref{eq:relavance_score}. 
To assess the benefit of combining global and local aggregated features in optimizing the loss, we compare the performance of FineLIP with a variant using only local features, as shown in \cref{tab:ablation1}D.
The results indicate that retaining global embeddings enhances performance, especially on the more challenging \textit{DOCCI} dataset (0.760 \vs 0.771 for I2T R@1 and 0.773 \vs 0.795 for T2I R@1). 
Thus, preserving features of varying granularity within the same modality improves representation learning. 



\section{Conclusion and Future Work}
\label{sec:conclusion_future_work}

This paper introduces FineLIP, a framework designed to improve long-caption image-text alignment and enhance cross-modal fine-grained understanding. FineLIP refines local image and text tokens, dynamically aggregating them into representations with improved information density. These aggregated tokens are then processed via a token-to-token module for precise cross-modal alignment. Our evaluations on long-caption retrieval and long-text-to-image generation benchmarks validate its effectiveness.

FineLIP is primarily optimized for long-text scenarios, and its performance on short-caption retrieval and classification is less satisfactory, consistent with prior findings \cite{lotlip}. Addressing this limitation by developing a more versatile model, such as one trained on both long and short captions \cite{longclip, tulip}, presents a promising direction for future work. \par
{
    \small
    \bibliographystyle{ieeenat_fullname}
    \bibliography{main}

\begin{thebibliography}{38}
\providecommand{\natexlab}[1]{#1}
\providecommand{\url}[1]{\texttt{#1}}
\expandafter\ifx\csname urlstyle\endcsname\relax
  \providecommand{\doi}[1]{doi: #1}\else
  \providecommand{\doi}{doi: \begingroup \urlstyle{rm}\Url}\fi

\bibitem[Balntas et~al.(2016)Balntas, Riba, Ponsa, and Mikolajczyk]{triplet}
Vassileios Balntas, Edgar Riba, Daniel Ponsa, and Krystian Mikolajczyk.
\newblock Learning local feature descriptors with triplets and shallow convolutional neural networks.
\newblock In \emph{Bmvc}, page~3, 2016.

\bibitem[Bica et~al.(2024)Bica, Ilic, Bauer, Erdogan, Bo{\v{s}}njak, Kaplanis, Gritsenko, Minderer, Blundell, Pascanu, and Mitrovic]{sparc}
Ioana Bica, Anastasija Ilic, Matthias Bauer, Goker Erdogan, Matko Bo{\v{s}}njak, Christos Kaplanis, Alexey~A. Gritsenko, Matthias Minderer, Charles Blundell, Razvan Pascanu, and Jovana Mitrovic.
\newblock Improving fine-grained understanding in image-text pre-training.
\newblock In \emph{Forty-first International Conference on Machine Learning}, 2024.

\bibitem[Bolya et~al.(2023)Bolya, Fu, Dai, Zhang, Feichtenhofer, and Hoffman]{tome}
Daniel Bolya, Cheng-Yang Fu, Xiaoliang Dai, Peizhao Zhang, Christoph Feichtenhofer, and Judy Hoffman.
\newblock Token merging: Your vit but faster.
\newblock In \emph{The Eleventh International Conference on Learning Representations}, 2023.

\bibitem[Chen et~al.(2023)Chen, Li, Dong, Zhang, He, Wang, Zhao, and Lin]{sharegpt}
Lin Chen, Jinsong Li, Xiaoyi Dong, Pan Zhang, Conghui He, Jiaqi Wang, Feng Zhao, and Dahua Lin.
\newblock Sharegpt4v: Improving large multi-modal models with better captions.
\newblock \emph{CoRR}, abs/2311.12793, 2023.

\bibitem[Dosovitskiy et~al.(2021)Dosovitskiy, Beyer, Kolesnikov, Weissenborn, Zhai, Unterthiner, Dehghani, Minderer, Heigold, Gelly, Uszkoreit, and Houlsby]{vit}
Alexey Dosovitskiy, Lucas Beyer, Alexander Kolesnikov, Dirk Weissenborn, Xiaohua Zhai, Thomas Unterthiner, Mostafa Dehghani, Matthias Minderer, Georg Heigold, Sylvain Gelly, Jakob Uszkoreit, and Neil Houlsby.
\newblock An image is worth 16x16 words: Transformers for image recognition at scale.
\newblock In \emph{International Conference on Learning Representations}, 2021.

\bibitem[Fu et~al.(2024)Fu, Zhang, Xia, and Mao]{laps}
Zheren Fu, Lei Zhang, Hou Xia, and Zhendong Mao.
\newblock Linguistic-aware patch slimming framework for fine-grained cross-modal alignment.
\newblock In \emph{Proceedings of the IEEE/CVF Conference on Computer Vision and Pattern Recognition (CVPR)}, pages 26307--26316, 2024.

\bibitem[Gadre et~al.(2023)Gadre, Ilharco, Fang, Hayase, Smyrnis, Nguyen, Marten, Wortsman, Ghosh, Zhang, Orgad, Entezari, Daras, Pratt, Ramanujan, Bitton, Marathe, Mussmann, Vencu, Cherti, Krishna, Koh, Saukh, Ratner, Song, Hajishirzi, Farhadi, Beaumont, Oh, Dimakis, Jitsev, Carmon, Shankar, and Schmidt]{datacomp}
Samir~Yitzhak Gadre, Gabriel Ilharco, Alex Fang, Jonathan Hayase, Georgios Smyrnis, Thao Nguyen, Ryan Marten, Mitchell Wortsman, Dhruba Ghosh, Jieyu Zhang, Eyal Orgad, Rahim Entezari, Giannis Daras, Sarah Pratt, Vivek Ramanujan, Yonatan Bitton, Kalyani Marathe, Stephen Mussmann, Richard Vencu, Mehdi Cherti, Ranjay Krishna, Pang Wei~W Koh, Olga Saukh, Alexander~J Ratner, Shuran Song, Hannaneh Hajishirzi, Ali Farhadi, Romain Beaumont, Sewoong Oh, Alex Dimakis, Jenia Jitsev, Yair Carmon, Vaishaal Shankar, and Ludwig Schmidt.
\newblock Datacomp: In search of the next generation of multimodal datasets.
\newblock In \emph{Advances in Neural Information Processing Systems}, pages 27092--27112. Curran Associates, Inc., 2023.

\bibitem[Gao et~al.(2022)Gao, Liu, Xu, Zhang, Li, Ji, and Shen]{pyramidclip}
Yuting Gao, Jinfeng Liu, Zihan Xu, Jun Zhang, Ke Li, Rongrong Ji, and Chunhua Shen.
\newblock Pyramidclip: Hierarchical feature alignment for vision-language model pretraining.
\newblock \emph{Advances in neural information processing systems}, 35:\penalty0 35959--35970, 2022.

\bibitem[Garg et~al.(2024)Garg, Burns, Ayan, Bitton, Montgomery, Onoe, Bunner, Krishna, Baldridge, and Soricut]{iiw}
Roopal Garg, Andrea Burns, Burcu~Karagol Ayan, Yonatan Bitton, Ceslee Montgomery, Yasumasa Onoe, Andrew Bunner, Ranjay Krishna, Jason Baldridge, and Radu Soricut.
\newblock Imageinwords: Unlocking hyper-detailed image descriptions.
\newblock \emph{arXiv preprint arXiv:2405.02793}, 2024.

\bibitem[He et~al.(2020)He, Fan, Wu, Xie, and Girshick]{moco}
Kaiming He, Haoqi Fan, Yuxin Wu, Saining Xie, and Ross Girshick.
\newblock Momentum contrast for unsupervised visual representation learning.
\newblock In \emph{Proceedings of the IEEE/CVF conference on computer vision and pattern recognition}, pages 9729--9738, 2020.

\bibitem[Hendrycks and Gimpel(2017)]{gelu}
Dan Hendrycks and Kevin Gimpel.
\newblock Bridging nonlinearities and stochastic regularizers with gaussian error linear units, 2017.

\bibitem[Heusel et~al.(2017)Heusel, Ramsauer, Unterthiner, Nessler, and Hochreiter]{fid}
Martin Heusel, Hubert Ramsauer, Thomas Unterthiner, Bernhard Nessler, and Sepp Hochreiter.
\newblock Gans trained by a two time-scale update rule converge to a local nash equilibrium.
\newblock \emph{Advances in neural information processing systems}, 30, 2017.

\bibitem[Jia et~al.(2021)Jia, Yang, Xia, Chen, Parekh, Pham, Le, Sung, Li, and Duerig]{align}
Chao Jia, Yinfei Yang, Ye Xia, Yi{-}Ting Chen, Zarana Parekh, Hieu Pham, Quoc~V. Le, Yun{-}Hsuan Sung, Zhen Li, and Tom Duerig.
\newblock Scaling up visual and vision-language representation learning with noisy text supervision.
\newblock \emph{CoRR}, abs/2102.05918, 2021.

\bibitem[Krishna et~al.(2017)Krishna, Zhu, Groth, Johnson, Hata, Kravitz, Chen, Kalantidis, Li, Shamma, et~al.]{visual_genome}
Ranjay Krishna, Yuke Zhu, Oliver Groth, Justin Johnson, Kenji Hata, Joshua Kravitz, Stephanie Chen, Yannis Kalantidis, Li-Jia Li, David~A Shamma, et~al.
\newblock Visual genome: Connecting language and vision using crowdsourced dense image annotations.
\newblock \emph{International journal of computer vision}, 123:\penalty0 32--73, 2017.

\bibitem[Lin et~al.(2014)Lin, Maire, Belongie, Hays, Perona, Ramanan, Doll{\'a}r, and Zitnick]{coco}
Tsung-Yi Lin, Michael Maire, Serge Belongie, James Hays, Pietro Perona, Deva Ramanan, Piotr Doll{\'a}r, and C~Lawrence Zitnick.
\newblock Microsoft coco: Common objects in context.
\newblock In \emph{Computer Vision--ECCV 2014: 13th European Conference, Zurich, Switzerland, September 6-12, 2014, Proceedings, Part V 13}, pages 740--755. Springer, 2014.

\bibitem[Liu et~al.(2024)Liu, Li, Wu, and Lee]{llava}
Haotian Liu, Chunyuan Li, Qingyang Wu, and Yong~Jae Lee.
\newblock Visual instruction tuning.
\newblock \emph{Advances in neural information processing systems}, 36, 2024.

\bibitem[Meng et~al.(2021)Meng, Li, Chen, Lan, Wu, Jiang, and Lim]{adaViT}
Lingchen Meng, Hengduo Li, Bor{-}Chun Chen, Shiyi Lan, Zuxuan Wu, Yu{-}Gang Jiang, and Ser{-}Nam Lim.
\newblock Adavit: Adaptive vision transformers for efficient image recognition.
\newblock \emph{CoRR}, abs/2111.15668, 2021.

\bibitem[Miranda et~al.(2024)Miranda, Salaberria, Agirre, and Azkune]{t2i-explain-1}
Imanol Miranda, Ander Salaberria, Eneko Agirre, and Gorka Azkune.
\newblock Bivlc: Extending vision-language compositionality evaluation with text-to-image retrieval.
\newblock \emph{arXiv preprint arXiv:2406.09952}, 2024.

\bibitem[Najdenkoska et~al.(2024)Najdenkoska, Derakhshani, Asano, van Noord, Worring, and Snoek]{tulip}
Ivona Najdenkoska, Mohammad~Mahdi Derakhshani, Yuki~M Asano, Nanne van Noord, Marcel Worring, and Cees~GM Snoek.
\newblock Tulip: Token-length upgraded clip.
\newblock \emph{arXiv preprint arXiv:2410.10034}, 2024.

\bibitem[Onoe et~al.(2024)Onoe, Rane, Berger, Bitton, Cho, Garg, Ku, Parekh, Pont-Tuset, Tanzer, et~al.]{docci}
Yasumasa Onoe, Sunayana Rane, Zachary Berger, Yonatan Bitton, Jaemin Cho, Roopal Garg, Alexander Ku, Zarana Parekh, Jordi Pont-Tuset, Garrett Tanzer, et~al.
\newblock Docci: Descriptions of connected and contrasting images.
\newblock \emph{arXiv preprint arXiv:2404.19753}, 2024.

\bibitem[OpenAI(2021)]{openai2021clip}
OpenAI.
\newblock Clip: Contrastive language–image pretraining.
\newblock \url{https://github.com/openai/CLIP}, 2021.
\newblock Accessed: [Date you accessed].

\bibitem[OpenAI(2023)]{gpt4}
OpenAI.
\newblock {GPT-4} technical report.
\newblock \emph{CoRR}, abs/2303.08774, 2023.

\bibitem[Parcalabescu et~al.(2021)Parcalabescu, Cafagna, Muradjan, Frank, Calixto, and Gatt]{valse}
Letitia Parcalabescu, Michele Cafagna, Lilitta Muradjan, Anette Frank, Iacer Calixto, and Albert Gatt.
\newblock Valse: A task-independent benchmark for vision and language models centered on linguistic phenomena.
\newblock \emph{arXiv preprint arXiv:2112.07566}, 2021.

\bibitem[Radford et~al.(2021)Radford, Kim, Hallacy, Ramesh, Goh, Agarwal, Sastry, Askell, Mishkin, Clark, et~al.]{clip}
Alec Radford, Jong~Wook Kim, Chris Hallacy, Aditya Ramesh, Gabriel Goh, Sandhini Agarwal, Girish Sastry, Amanda Askell, Pamela Mishkin, Jack Clark, et~al.
\newblock Learning transferable visual models from natural language supervision.
\newblock In \emph{International conference on machine learning}, pages 8748--8763. PMLR, 2021.

\bibitem[Ranasinghe et~al.(2023)Ranasinghe, McKinzie, Ravi, Yang, Toshev, and Shlens]{perceptual}
Kanchana Ranasinghe, Brandon McKinzie, Sachin Ravi, Yinfei Yang, Alexander~T Toshev, and Jonathon Shlens.
\newblock Perceptual grouping in vision-language models, 2023.

\bibitem[Rao et~al.(2021)Rao, Zhao, Liu, Lu, Zhou, and Hsieh]{Dynamicvit}
Yongming Rao, Wenliang Zhao, Benlin Liu, Jiwen Lu, Jie Zhou, and Cho{-}Jui Hsieh.
\newblock Dynamicvit: Efficient vision transformers with dynamic token sparsification.
\newblock \emph{CoRR}, abs/2106.02034, 2021.

\bibitem[Ray et~al.(2024)Ray, Radenovic, Dubey, Plummer, Krishna, and Saenko]{t2i-explain-3}
Arijit Ray, Filip Radenovic, Abhimanyu Dubey, Bryan Plummer, Ranjay Krishna, and Kate Saenko.
\newblock Cola: A benchmark for compositional text-to-image retrieval.
\newblock \emph{Advances in Neural Information Processing Systems}, 36, 2024.

\bibitem[Rombach et~al.(2022)Rombach, Blattmann, Lorenz, Esser, and Ommer]{SDXL}
Robin Rombach, Andreas Blattmann, Dominik Lorenz, Patrick Esser, and Bj{\"o}rn Ommer.
\newblock High-resolution image synthesis with latent diffusion models.
\newblock In \emph{Proceedings of the IEEE/CVF conference on computer vision and pattern recognition}, pages 10684--10695, 2022.

\bibitem[Thrush et~al.(2022)Thrush, Jiang, Bartolo, Singh, Williams, Kiela, and Ross]{t2i-explain-2}
Tristan Thrush, Ryan Jiang, Max Bartolo, Amanpreet Singh, Adina Williams, Douwe Kiela, and Candace Ross.
\newblock Winoground: Probing vision and language models for visio-linguistic compositionality.
\newblock In \emph{Proceedings of the IEEE/CVF Conference on Computer Vision and Pattern Recognition}, pages 5238--5248, 2022.

\bibitem[Touvron et~al.(2023)Touvron, Martin, Stone, Albert, Almahairi, Babaei, Bashlykov, Batra, Bhargava, Bhosale, et~al.]{llama}
Hugo Touvron, Louis Martin, Kevin Stone, Peter Albert, Amjad Almahairi, Yasmine Babaei, Nikolay Bashlykov, Soumya Batra, Prajjwal Bhargava, Shruti Bhosale, et~al.
\newblock Llama 2: Open foundation and fine-tuned chat models.
\newblock \emph{arXiv preprint arXiv:2307.09288}, 2023.

\bibitem[Urbanek et~al.(2024)Urbanek, Bordes, Astolfi, Williamson, Sharma, and Romero-Soriano]{dci}
Jack Urbanek, Florian Bordes, Pietro Astolfi, Mary Williamson, Vasu Sharma, and Adriana Romero-Soriano.
\newblock A picture is worth more than 77 text tokens: Evaluating clip-style models on dense captions.
\newblock In \emph{Proceedings of the IEEE/CVF Conference on Computer Vision and Pattern Recognition}, pages 26700--26709, 2024.

\bibitem[Wu et~al.(2024)Wu, Zheng, Ma, Lu, Guo, Zhang, Chen, Guo, Shen, and Zha]{lotlip}
Wei Wu, Kecheng Zheng, Shuailei Ma, Fan Lu, Yuxin Guo, Yifei Zhang, Wei Chen, Qingpei Guo, Yujun Shen, and Zheng-Jun Zha.
\newblock Lo{TLIP}: Improving language-image pre-training for long text understanding.
\newblock In \emph{The Thirty-eighth Annual Conference on Neural Information Processing Systems}, 2024.

\bibitem[Yao et~al.(2022)Yao, Huang, Hou, Lu, Niu, Xu, Liang, Li, Jiang, and Xu]{filip}
Lewei Yao, Runhui Huang, Lu Hou, Guansong Lu, Minzhe Niu, Hang Xu, Xiaodan Liang, Zhenguo Li, Xin Jiang, and Chunjing Xu.
\newblock {FILIP}: Fine-grained interactive language-image pre-training.
\newblock In \emph{International Conference on Learning Representations}, 2022.

\bibitem[Young et~al.(2014)Young, Lai, Hodosh, and Hockenmaier]{flickr30k}
Peter Young, Alice Lai, Micah Hodosh, and Julia Hockenmaier.
\newblock From image descriptions to visual denotations: New similarity metrics for semantic inference over event descriptions.
\newblock \emph{Transactions of the Association for Computational Linguistics}, 2:\penalty0 67--78, 2014.

\bibitem[Zhai et~al.(2022)Zhai, Wang, Mustafa, Steiner, Keysers, Kolesnikov, and Beyer]{lit}
Xiaohua Zhai, Xiao Wang, Basil Mustafa, Andreas Steiner, Daniel Keysers, Alexander Kolesnikov, and Lucas Beyer.
\newblock Lit: Zero-shot transfer with locked-image text tuning.
\newblock In \emph{Proceedings of the IEEE/CVF conference on computer vision and pattern recognition}, pages 18123--18133, 2022.

\bibitem[Zhang et~al.(2024)Zhang, Zhang, Dong, Zang, and Wang]{longclip}
Beichen Zhang, Pan Zhang, Xiaoyi Dong, Yuhang Zang, and Jiaqi Wang.
\newblock Long-clip: Unlocking the long-text capability of clip.
\newblock \emph{arXiv preprint arXiv:2403.15378}, 2024.

\bibitem[Zheng et~al.(2025)Zheng, Zhang, Wu, Lu, Ma, Jin, Chen, and Shen]{dreamlip}
Kecheng Zheng, Yifei Zhang, Wei Wu, Fan Lu, Shuailei Ma, Xin Jin, Wei Chen, and Yujun Shen.
\newblock Dreamlip: Language-image pre-training with long captions.
\newblock In \emph{European Conference on Computer Vision}, pages 73--90. Springer, 2025.

\bibitem[Zong et~al.(2022)Zong, Li, Song, Wang, Qiao, Leng, and Liu]{tsm}
Zhuofan Zong, Kunchang Li, Guanglu Song, Yali Wang, Yu Qiao, Biao Leng, and Yu Liu.
\newblock Self-slimmed vision transformer.
\newblock In \emph{Computer Vision -- ECCV 2022}, pages 432--448, Cham, 2022. Springer Nature Switzerland.

\end{thebibliography}
}

\clearpage
\setcounter{page}{1}
\maketitlesupplementary

\section{Details of long-text-to-image generation}

In \cref{fig:generation}, the captions used to generate images from top to down are listed.

\begin{enumerate}[label=(\arabic*)]
    \item \textit{A rather low aerial view from an airplane at a football field with bleachers on each side. The football field is in the center of the frame and angled from bottom left to upper right. The image is blurry. The field is green with white football striping and markings. The goal posts are on each end. There is a red track and deck around the field. The bleachers on the left side have a control box on its top center. Behind the football to the right is a soccer field. To the left of the soccer field is a baseball field. The forefront is a city street void of traffic. The flat top of a commercial building is in the bottom right corner of the frame. Across the top of the frame in the background is a city neighborhood that is speckled with house rooftops and trees. }
    \item \textit{An outdoor close-up view looking up at a painting on a tall worn down gray colored wall that has cracks and dark markings spread throughout its surface. Towards the bottom of the painting is a large flower with dark pink petals and an orange stigma, below the flower are faded black painted block letters placed side by side that have a white faded blob on them. Above the flower and to its sides are small scattered out yellow painted stars that are circle shaped, and about a foot above the flower is a large painting of a beige colored dragonfly with a pink head and white wings. Above the gray colored wall is a silver metal caged fence, the clear light blue sky can be seen through the fence and above it.}
    \item \textit{A side view of a white Chevrolet Silverado pickup truck parked on the street outside of a two story home that's painted a light blue color on the outside with white colors around the windows and corners across the home. The white Chevrolet truck has black rims and has slight damage and scratches along the visible body. The truck blocks the full view of the yard outside the home, but two large trees with numerous leaves are visible on the left and right of the view, with green grass and dry patches visible below. At the bottom of the view, healthy green grass can be seen in the bottom right, while green plants are visible to the left growing out of the dirt. The plants and grass are separated by a black plastic divider located in the bottom middle portion of the view. Numerous shadows are cast as the view is illuminated by sunlight, At the bottom of the view, the street is covered in shadows from nearby trees, branches, and leaves. The white Chevrolet pickup truck is partially bright from sunlight, but also a shadow is visible from the driver side mirror extending toward the bottom left of the view. In the background behind the truck, the tree to the left in the yard is bright from sunlight, while the tree to the right is slightly darker and cast in partial shadows.}
    \item \textit{An indoor medium close-up front view of a doorway that has a white wooden frame and a white door that is currently swung open to the left side. The doorway leads into a lit up bathroom that has a light colored wooden floor made up of wooden panels placed side by side. On the right side of the bathroom is a white countertop that consists of one sink, and dark brown drawers and cabinet doors below it. There is a white bath mat placed on the floor, in front of the countertop. To the left side of the countertop is a white wall that has a rectangular shaped mirror mounted to it that is positioned vertically. Behind the white wall is a partial view of a white toilet that is pointed out towards the left, and to the left of the toilet are several white towels hanging from a towel rack that is mounted to the wall}.

\end{enumerate}
Combining the \cref{fig:generation} and the provided captions here, we can have the following observations:
\begin{enumerate}[label=(\arabic*)]
    \item For the first example, LAPS and our FineLIP can capture the keywords \textit{``red track and deck around the field"}, while the others cannot. 
    \item In the second caption, the keyword \textit{``dragonfly”} is only caught by Baseline and FineLIP, as shown at the bottom left of noteworthy flower.
    \item Regarding the \textit{``white truck"} in the third caption, CLIP(L/14) and SPARC fail, Baseline seems to just give a partial view, while LAPS and FineLIP presents the whole truck with higher quality.
    \item Finally, in the last example, the generated image from FineLIP accurately reflects the keywords \textit{``cabinet"} and \textit{``toilet"} as specified in the caption, while other models fail to do so.
\end{enumerate}

\vspace{1em}

\begin{table*}[ht]
    \centering
    \resizebox{2\columnwidth}{!}{\ShortCaptionA}
    \caption{Short caption cross-modal retrieval on \textit{Flickr30k} and \textit{COCO}. Best result is in bold.}
    \label{tab:short_caption}
\end{table*}

Although these visualized examples highlight the effectiveness of FineLIP compared to other state-of-the-art methods, many details (such as color, attributes, and locations) are still missing in the generated images. Furthermore, the gap between the generated images and the ground-truth images remains substantial. It would be both interesting and meaningful to explore how to effectively leverage long captions in the end-to-end training of text-to-image generation models.

\section{Long caption datasets details and additional results}\label{sup:eval}
\begin{table}[ht]
    \centering
    \resizebox{\columnwidth}{!}{\additionalExp}
    \caption{Zero-shot cross-modal retrieval results on additional long caption datasets}
    \label{tab:long_caption_2}
\end{table}
\textbf{\textit{Urban1k} dataset} contains 1,000 (image, long caption) pairs. 
The images selected from the Visual Genome dataset \cite{visual_genome} depict busy urban scenes, with captions generated by GPT-4V \cite{gpt4} averaging 101 words. 
These detailed captions describe attributes such as object types, colors, and spatial relationships, making this dataset particularly suitable for evaluating zero-shot long caption cross-modal retrieval.
\noindent\textbf{\textit{DOCCI} dataset} offers long, human-annotated captions for 15,000 images, with an average description length of 136 words. 
\textit{DOCCI}'s descriptions are highly detailed, covering complex visual challenges like spatial relationships between objects, counting, and text rendering. We utilize the official 5k test split for evaluation.

For long-caption image-text alignment, we added additional results on long-caption retrieval datasets, \textit{Long-DCI} \cite{dci} and \textit{IIW} \cite{iiw} (see \cref{tab:long_caption_2}). 

\section{Evaluation on zero-shot short caption cross-modal retrieval and classification}

\begin{figure*}[ht]
    \centering
    \includegraphics[width=\textwidth]{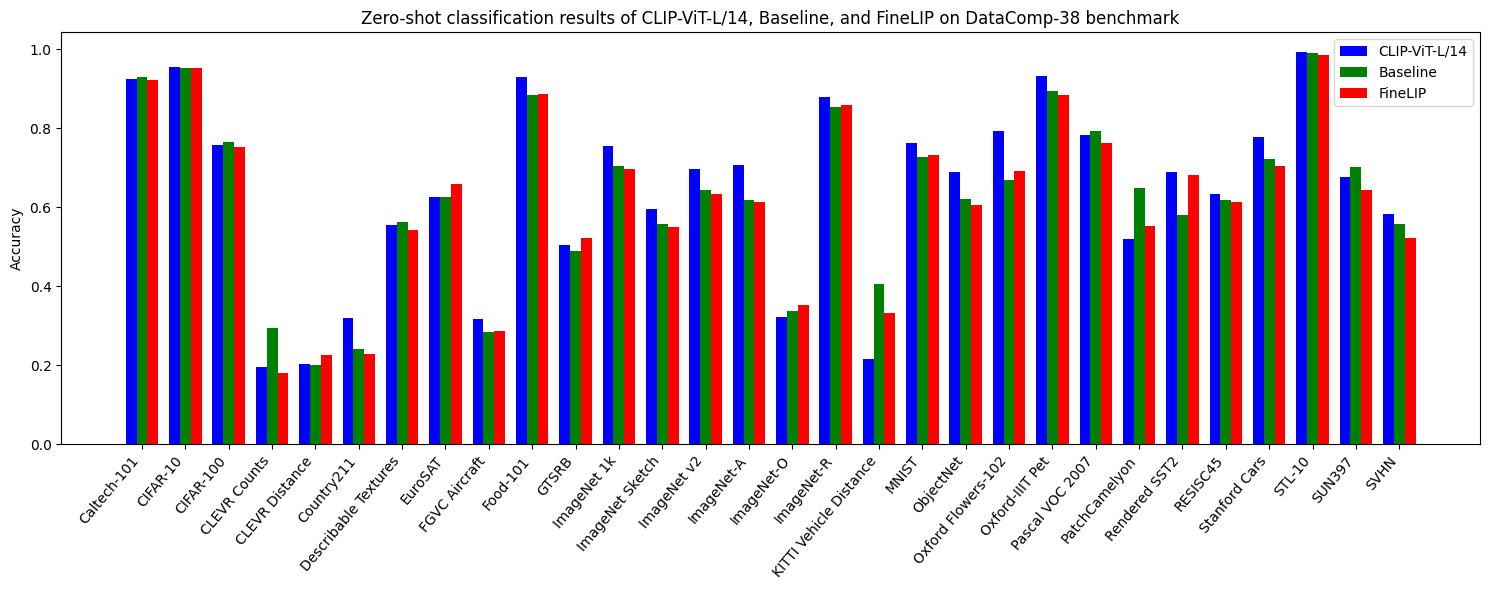}
    \caption{Zero-shot Classification on \textit{DataComp-38}}
    \label{fig:zsc}
\end{figure*}

Our approach, FineLIP, is specifically designed to handle long captions, so training and evaluation are conducted exclusively with long caption data. 
Interestingly, we find that FineLIP also gives a significant performance boost on tasks involving short captions, even though short captions are not explicitly used in training and are not the primary focus of this paper.

To have the comprehensive evaluation,  we adopt the widely used zero-shot cross-modal retrieval benchmarks for short captions: \textit{Flickr30k} \cite{flickr30k} and \textit{COCO2017} \cite{coco}. In \textit{COCO2017}, we use the 5k validation set, while for \textit{Flickr30k}, we employ the entire 30k dataset following the standard practice. 
In \cref{tab:short_caption}, the Pretrained-CLIP\footnote{\href{https://openaipublic.azureedge.net/clip/models/5806e77cd80f8b59890b7e101eabd078d9fb84e6937f9e85e4ecb61988df416f/ViT-B-16.pt}{\nolinkurl{CLIP-ViT-B/16}}}\footnote{\href{https://openaipublic.azureedge.net/clip/models/b8cca3fd41ae0c99ba7e8951adf17d267cdb84cd88be6f7c2e0eca1737a03836/ViT-L-14.pt}{\nolinkurl{CLIP-ViT-L/14}}} models refer to those released by OpenAI, trained on 400 million short-text image pairs from web sources, and demonstrate strong performance on short caption datasets. After fine-tuning on \textit{ShareGPT4}, the long caption dataset, with traditional contrastive learning (Baseline), substantial improvements are observed across all metrics, benchmarks, and model sizes. FineLIP, which incorporates aggregation and alignment modules for more fine-grained cross-modal alignment, significantly outperforms the baseline in image-to-text retrieval and achieves comparable performance in text-to-image retrieval. TULIP, trained with long and short captions, delivers results that fall between these two approaches. 

The results indicate that training with longer captions enhances the model's generalization ability, improving performance on tasks involving short captions. This promising finding highlights the advantages of leveraging longer, more detailed captions and the importance of fine-grained alignment, even for short caption tasks. However, we also observe that FineLIP achieves text-to-image results comparable to the Baseline while providing significant gains in image-to-text retrieval. This pattern, consistent with findings from previous studies on cross-modal retrieval \cite{t2i-explain-1,t2i-explain-2,t2i-explain-3}, needs further investigation.

For zero-shot classification, we evaluate on \textit{DataComp-38} \cite{datacomp} (see \cref{fig:zsc}), with Pretrained-CLIP, Baseline, and FineLIP achieving average accuracies of 0.643, 0.629, and 0.619, respectively.


\begin{table}[ht]
    \centering
    \resizebox{\columnwidth}{!}{\CLIMAbalation}
    \caption{Effect of CLIM in cross-modality alignment}
    \label{tab:clim}
\end{table}

\section{Importance of CLIM}
We conducted an ablation study on the CLIM block. Keeping ATRM fixed, we tested two cross-modality alignment methods. When using baseline contrastive loss, which is generally designed to perform on global features (the feature of [CLS] token for image and the feature of [EOS] token for text), no improvement is observed as ATRM only refines local tokens. ATRM + FILIP  (\cref{tab:clim}) underperformed our solution FineLIP (ATRM + CLIM), likely due to FILIP’s training instability, as noted in its original paper \cite{filip} and SPARC \cite{sparc}.

\section{Visualization of long caption image-text retrieval}

\begin{figure*}[t]
  \centering
 \includegraphics[width=\textwidth]{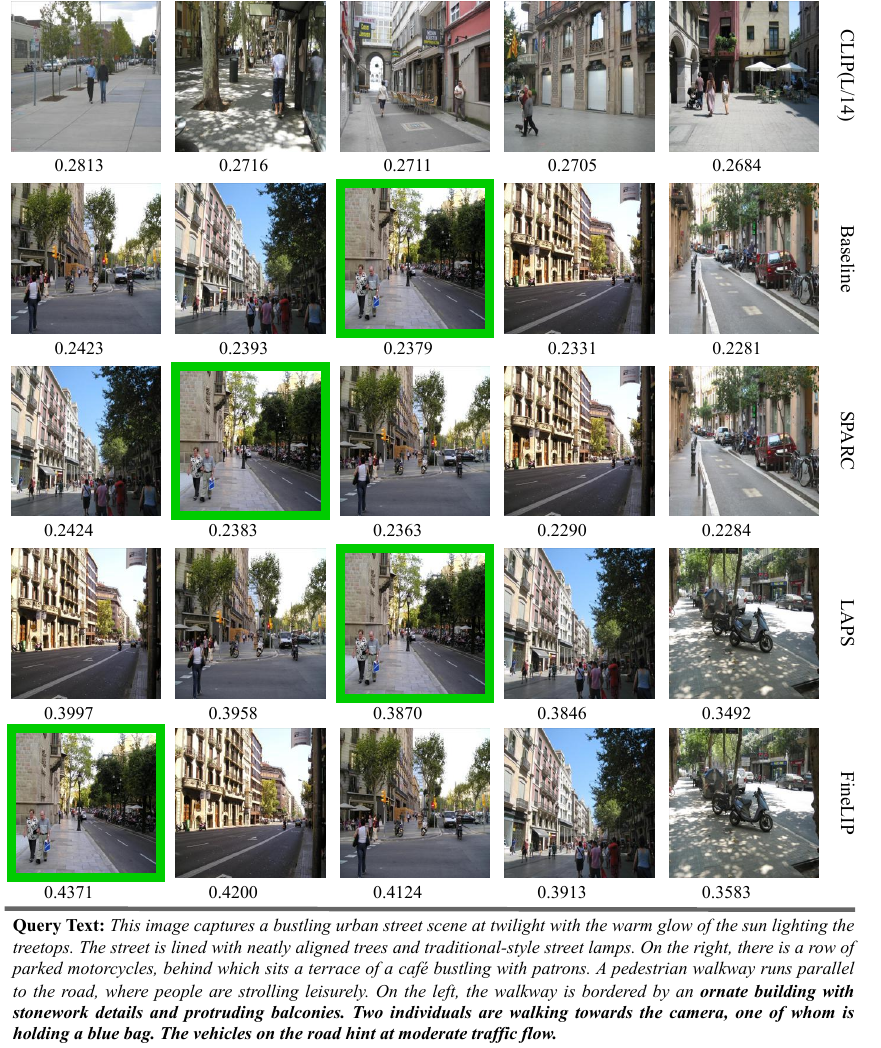}
   \caption{Top-5 text-to-image retrieval results on \textit{Urban1k} dataset \cite{longclip} for L/14 variants of CLIP, Baseline, SPARC \cite{sparc}, LAPS \cite{laps} and FineLIP (Ours), with retrieval scores. The correct retrieved images are marked with green boxes. CLIP ignores the caption in bold due to the 77 token limit.}
   \label{fig:t2i}
\end{figure*}

\begin{figure*}[ht]
  \centering
   \includegraphics[width=\textwidth,height=0.99\textheight]{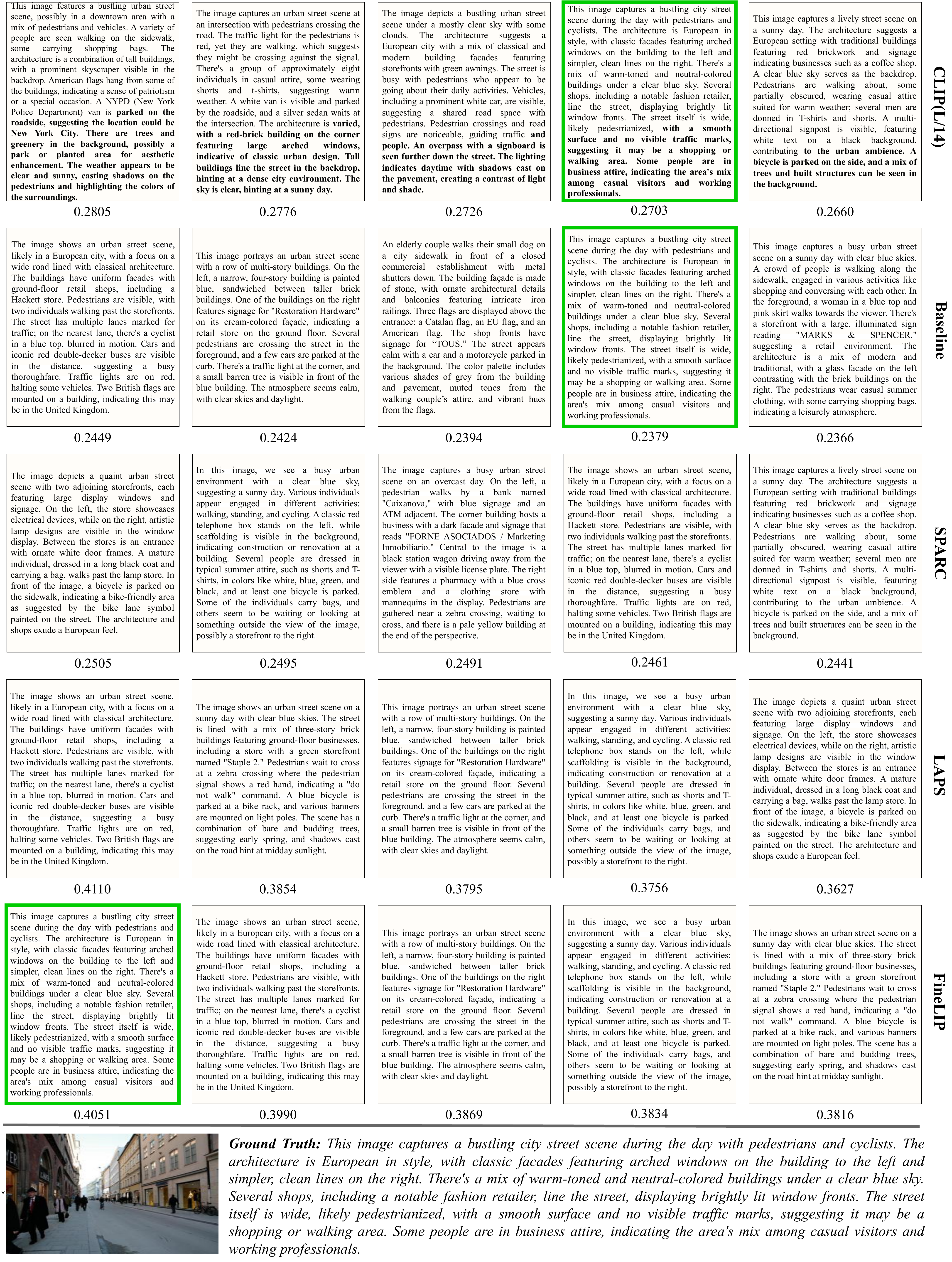}
   \caption{Top-5 image-to-text retrieval results on \textit{Urban1k} dataset \cite{longclip} for L/14 variants of CLIP, Baseline, SPARC \cite{sparc}, LAPS \cite{laps} and FineLIP (Ours), with retrieval scores. The correct retrieved texts are marked with green boxes. CLIP ignores the text in bold due to the 77-token limit. Zoom in for better visualization.}
   \label{fig:i2t}
\end{figure*}

\cref{fig:t2i} presents the Text-to-Image (T2I) retrieval results for various models on \textit{Urban1k} dataset. The grid shows the top-5 retrieved images for a given text query, ranked by the similarity score between the text and image features. The higher the score, the more relevant the image is to the query. 
The ground truth image (GT) does not appear within the top 5 retrieval results for CLIP (L/14). For the Baseline, SPARC, and LAPS, the top 1 retrieval results are not the GT but rather images that are highly similar to it. In contrast, FineLIP retrieves the GT in the top position, demonstrating its superior ability to capture fine-grained details and to distinguish among highly similar images. 

\cref{fig:i2t} illustrates the Image-to-Text (I2T) retrieval results on the \textit{Urban1k} dataset. The figure shows the top-5 retrieved captions for a given image, ranked by similarity scores. 
FineLIP manages to retrieve the ground truth caption in the top position, while others models fail. In particular, for models like SPARC and LAPS, the correct caption is not even within the top-5 results. This example shows that FineLIP is capable of extracting discriminative fine-grained textual features even when the text input is long.

These visual results emphasize FineLIP’s advantage in enabling more precise and nuanced retrieval, especially for tasks involving complex text-to-image associations that demand detailed comprehension. FineLIP’s ability to leverage long captions and fine-grained alignment methods allows it to retrieve highly relevant content that other models may overlook. By extending the token limit and using aggregation before token-level alignment strategy, FineLIP achieves more accurate and contextually relevant retrieval than existing models.


\end{document}